\documentclass{article}

     \PassOptionsToPackage{numbers, compress}{natbib}

\usepackage[final]{neurips_2019}

\usepackage[utf8]{inputenc} 
\usepackage[T1]{fontenc}    
\usepackage{hyperref}       
\usepackage{url}            
\usepackage{booktabs}       
\usepackage{amsfonts}       
\usepackage{nicefrac}       
\usepackage{microtype}      

\usepackage{wrapfig}
\usepackage{sidecap}
\usepackage{amsmath, amssymb,bm}
\usepackage{amsfonts}
\usepackage{bbm}
\usepackage{graphicx}
\usepackage{subfigure}
\usepackage[linesnumbered,ruled]{algorithm2e}
\usepackage{algorithmic}
\usepackage{booktabs}

\newcommand{\tmpvect}[2]{\mbox{\boldmath $#1#2$}}
\newcommand{\vect}[1]{\mathpalette\tmpvect{#1}}

\makeatletter
\newcommand{\figcaption}[1]{\def\@captype{figure}\caption{#1}}
\newcommand{\tblcaption}[1]{\def\@captype{table}\caption{#1}}
\makeatother
\usepackage{xcolor}

\title{Spatially Aggregated Gaussian Processes\\
with Multivariate Areal Outputs}

\author{%
 Yusuke Tanaka\textsuperscript{1,3},
 Toshiyuki Tanaka\textsuperscript{3},
 Tomoharu Iwata\textsuperscript{2},
 Takeshi Kurashima\textsuperscript{1}, \and
 {\bf Maya Okawa\textsuperscript{1},
 Yasunori Akagi\textsuperscript{1},
 Hiroyuki Toda\textsuperscript{1}}\\
 \and
 \textsuperscript{1}{NTT Service Evolution Labs.},
 \textsuperscript{2}{NTT Communication Science Labs.},
 \textsuperscript{3}{Kyoto University}\\
 \and
 \texttt{\{yusuke.tanaka.rh,tomoharu.iwata.gy,takeshi.kurashima.uf,
 maya.ookawa.af,}\\
 \texttt{yasunori.akagi.cu,hiroyuki.toda.xb\}@hco.ntt.co.jp}, 
 \texttt{tt@i.kyoto-u.ac.jp}
}

\begin{document}

\maketitle

\begin{abstract}
We propose a probabilistic model
for inferring the multivariate function 
from multiple areal data sets with various granularities.
Here, the areal data are observed not at location points but at regions.
Existing regression-based models can only utilize
the sufficiently fine-grained auxiliary data sets
on the same domain (e.g., a city).
With the proposed model,
the functions for respective areal data sets
are assumed to be a multivariate dependent Gaussian process (GP)
that is modeled as a linear mixing of independent latent GPs.
Sharing of latent GPs across multiple areal data sets
allows us to effectively estimate
the spatial correlation for each areal data set;
moreover it can easily be extended to transfer learning across multiple domains.
To handle the multivariate areal data,
we design an observation model with a spatial aggregation process
for each areal data set,
which is an integral of the mixed GP
over the corresponding region.
By deriving the posterior GP,
we can predict the data value at any location point by
considering the spatial correlations
and the dependences between areal data sets, simultaneously.
Our experiments on real-world data sets
demonstrate that our model can
1) accurately refine coarse-grained areal data,
and 2) offer performance improvements by using the areal data sets from multiple domains.
\end{abstract}

\section{Introduction}
Governments and other organizations are now collecting
data from cities on items such as poverty rate, air pollution,
crime, energy consumption, and traffic flow.
These data play a crucial role in improving
the life quality of citizens
in many aspects including socio-economics~\cite{rupasinghaa:social,Smith:poverty},
public security~\cite{bogomolov:once,wang:crime},
public health~\cite{jerrett:spatial},
and urban planning~\cite{yuan:discovering}.
For instance, the spatial distribution of poverty
is helpful in identifying key regions that require intervention in a city;
it makes it easier to optimize resource allocation for remedial action.

\begin{wrapfigure}{r}{28mm}
 \vspace{-5pt}
 \begin{tabular}{c}
  \includegraphics[width=26mm, bb=0 0 1000 1000]{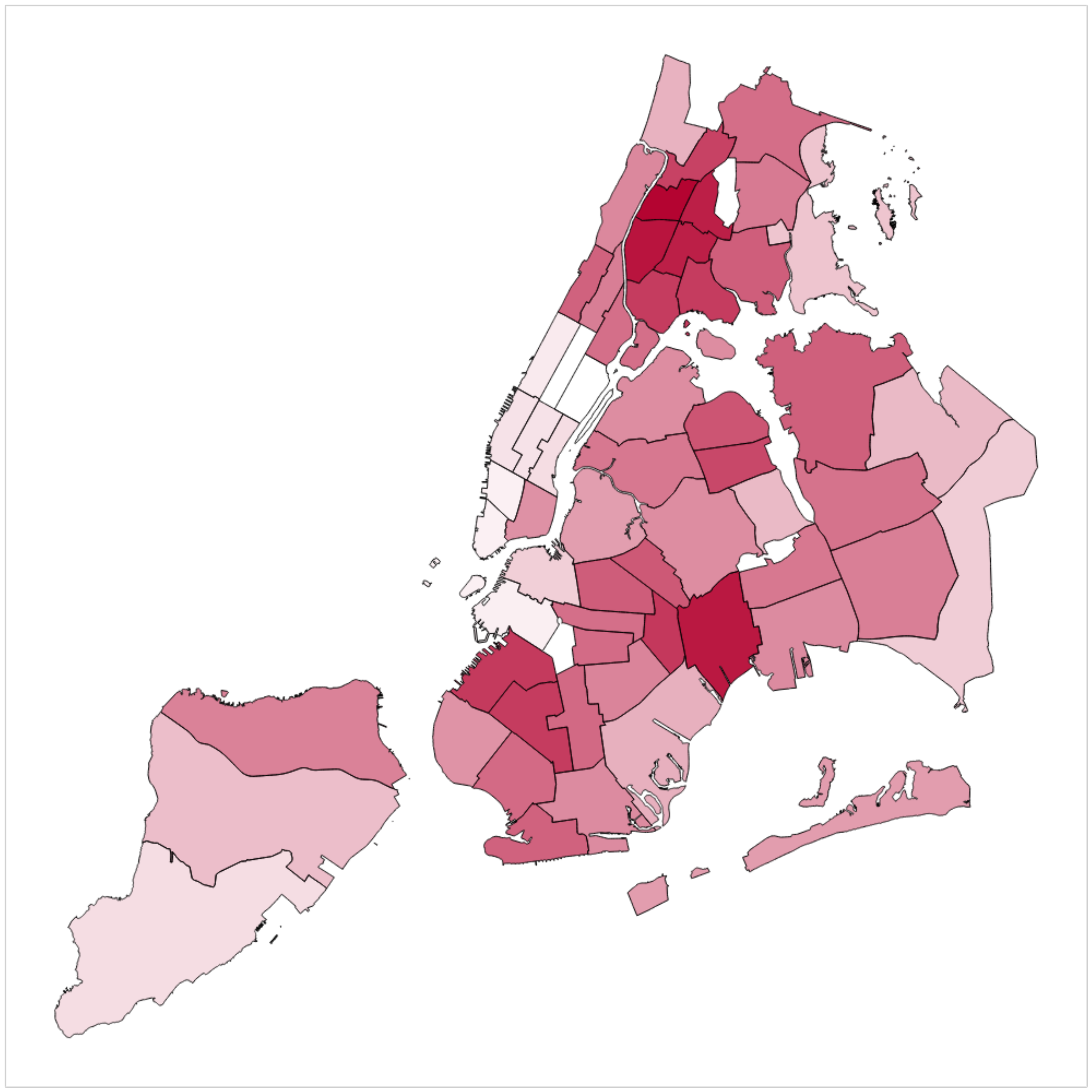}
 \end{tabular}
 \vspace{-10pt}
 \caption{Areal data}
 \label{fig:areal}
\end{wrapfigure}
In practice, the data collected from cities are often spatially aggregated,
e.g., averaged over a region;
thus only {\em areal data} are available;
observations are not associated with location points but with regions.
Figure~\ref{fig:areal} shows an example of areal data,
which is the distribution of poverty rate in New York City,
where darker hues represent regions with higher rates.
This poverty rate data set was actually obtained 
via household surveys taken over the whole city.
The survey results are aggregated over predefined regions~\cite{Smith:poverty}.
The problem addressed herein is
to infer the {\em function} from the areal data;
once we have the function we can predict data values
at any location point.
Solving this problem allows us to obtain
spatially-specific information about cities;
it is useful for finding key pin-point regions efficiently.

One promising approach to address this problem is
to utilize a wide variety of data sets from the same domain 
(e.g., a city).
Existing regression-based models learn relationships between target data and auxiliary data
sets~\cite{Law:variational,murakami:new,park:spatial,Smith:poverty,tanaka:refining}.
These models, however, assume that 
the auxiliary data sets have sufficiently fine spatial granularity
(e.g., 1~km $\times$ 1~km grid cells);
unfortunately, many areal data sets are actually associated
with large regions (e.g., zip code and police precinct).
These models cannot then make full use of the coarse-grained auxiliary data sets.
Another important drawback of all the prior works is that
their performance in refining the areal data is suspect
if we have only a few data sets available for the domain.

In this paper, we propose a probabilistic model,
called {\em Spatially Aggregated Gaussian Processes (SAGP)} herein after,
that can infer the multivariate function
from multiple areal data sets with various granularities.
In SAGP, the functions for the areal data sets
are assumed to be a multivariate dependent Gaussian process (GP)
that is modeled as a linear mixing of independent latent GPs.
The latent GPs are shared among all areal data sets in the target domain,
which is expected to effectively learn the spatial correlation for each data set
even if the number of observations in a data set is small; that is,
a data set associated with a coarse-grained region.
Since the areal data are identified by regions, not by location points, 
we introduce an observation model with the spatial aggregation process,
in which areal observations are assumed to be calculated
by integrating the mixed GP over the corresponding region;
then the covariance between regions is given by
the double integral of the covariance function 
over the corresponding pair of regions.
This allows us to accurately evaluate
the covariance between regions 
from a consideration of region shape.
Thus the proposal is very helpful 
if there are irregularly shaped regions (e.g., extremely elongated)
in the input data.

The mechanism adopted in SAGP for sharing latent processes
is also advantageous in that
it makes it straightforward to utilize data sets from multiple domains.
This allows our model to learn the spatial correlation for each data set
by sharing the latent GPs among all areal data sets from multiple domains;
SAGP remains applicable even if we have only a few data sets available for a single domain.

The inference of SAGP is based on a Bayesian inference procedure.
The model parameters can be estimated by maximizing the marginal likelihood,
in which all the GPs are analytically integrated out.
By deriving the posterior GP,
we can predict the data value at any location point
considering the spatial correlations
and the dependences between areal data sets, simultaneously.

The major contributions of this paper are as follows:
\begin{itemize}
 \setlength{\leftskip}{-0.5cm}
 \item We propose SAGP, a novel multivariate GP model
       that is defined by mixed latent GPs;
       it incorporates aggregation processes
       for handling multivariate areal data.
 \item We develop a parameter estimation procedure
       based on the marginal likelihood
       in which latent GPs are analytically integrated out.
       This is the first explicit derivation of
       the posterior GP given multivariate areal data;
       it allows the prediction of values at any location point.
 \item We conduct experiments on multiple real-world data sets
       from urban cities; the results show that our model
       can 1) accurately refine coarse-grained areal data,
       and 2) improve refinement performances
       by utilizing areal data sets from multiple cities.
\end{itemize}

\section{Related Work}
\label{sec:related}
Related works can be roughly categorized into two approaches:
1) regression-based model and 2) multivariate model.
The major difference between them is as follows:
Denoting ${\vect y}^{\rm t}$ and ${\vect y}$ as target data and auxiliary data, respectively,
the aim of the first approach is to design a conditional distribution $p({\vect y}^{\rm t} | {\vect y})$;
the second approach designs a joint distribution $p({\vect y}^{\rm t}, {\vect y})$.

{\bf Regression-based models.}
A related problem has been addressed in the spatial statistics community
under the name of {\em downscaling}, {\em spatial disaggregation},
{\em areal interpolation}, or {\em fine-scale modeling}~\cite{gotway:combining},
and this has attracted great interest in many disciplines
such as socio-economics~\cite{bogomolov:once,Smith:poverty},
agricultural economics~\cite{howitt:spatial,xavier:disaggregating},
epidemiology~\cite{sturrock:fine},
meteorology~\cite{wilby:guidelines,wotling:regionalization},
and geographical information systems (GIS)~\cite{goovaerts:combining}.
The problem of predicting point-referenced data
from areal observations is also related to
the {\em change of support problem}
in geostatistics~\cite{gotway:combining}.
Regression-based models have been developed for refining coarse-grained target data
via the use of multiple auxiliary data sets that have fine granularity
(e.g., 1~km $\times$ 1~km grid cells)~\cite{murakami:new,park:spatial}.
These models learn the regression coefficients for the auxiliary data sets
under the spatial aggregation constraints
that encourage consistency between fine- and coarse-grained target data.
The aggregation constraints have been incorporated via
{\em block kriging}~\cite{burgess:optimal}
or transformations of Gaussian process (GP)
priors~\cite{smith:transformations,smith:gaussian}.
There have been a number of advanced models
that offer a fully Bayesian inference~\cite{keil:downscaling,taylor:continuous,wilson:pointless}
or a variational inference~\cite{Law:variational} for model parameters.
The task addressed in these works is to refine the coarse-grained target data
on the assumption that the fine-grained auxiliary data are available;
however, the areal data available on a city are actually associated with
various geographical partitions (e.g., police precinct),
thus one might not be able to obtain the fine-grained auxiliary data.
In that case, these models cannot make full use of the auxiliary data sets
with various granularities, which contain the coarse-grained auxiliary data.

A GP-based model was recently proposed
for refining coarse-grained areal data by utilizing auxiliary data sets
with various granularities~\cite{tanaka:refining}.
In this model, GP regression is first applied to each auxiliary data set
for deriving a predictive distribution defined on the continuous space;
this conceptually corresponds to spatial interpolation.
By hierarchically incorporating the predictive distributions into the model,
the regression coefficients can be learned
on the basis of not only the strength of relationship with the target data
but also the level of spatial granularity.
A major disadvantage of this model is that
the spatial interpolation is separately conducted for each auxiliary data set,
which makes it difficult to accurately interpolate the coarse-grained auxiliary data
due to the data sparsity issue;
this model fails to fully use the coarse-grained data.

In addition, these regression-based models (e.g.,~\cite{Law:variational,park:spatial,tanaka:refining})
do not consider the spatial aggregation constraints
for the auxiliary data sets.
This is a critical issue 
in estimating the multivariate function
from multiple areal data sets,
the problem focused in this paper.

Different from the regression-based models,
we design a joint distribution that incorporates the spatial aggregation process
for all areal data sets (i.e., for both target and auxiliary data sets).
The proposed model infers the multivariate function 
while considering the spatial aggregation constraints for respective areal data sets.
This allows us to effectively utilize
all areal data sets with various granularities for the data refinement
even if some auxiliary data sets have coarse granularity.

{\bf Multivariate models.}
The proposed model builds closely upon recent studies in multivariate spatial modeling,
which model the joint distribution of multiple outputs.
Many geostatistics studies use the classical method of {\em co-kriging}
for predicting multivariate spatial data~\cite{myers:co-kriging};
this method is, however, problematic in that it is unclear how to define
cross-covariance functions that determine the dependences between data sets~\cite{gibbs:efficient}.
In the machine learning community,
there has been growing interest in multivariate GPs~\cite{carl:gaussian},
in which dependences between data sets are introduced via methodologies such as
process convolution~\cite{boyle:dependent,higdon:space},
latent factor modeling~\cite{luttinen:variational,teh:semiparametric},
and multi-task learning~\cite{bonilla:multi-task,micchelli:kernels}.
The linear model of coregionalization (LMC)
is one of the most widely-used approaches
for constructing a multivariate function;
the outputs are expressed as linear combinations of
independent latent functions~\cite{alvarez:kernels}.
The semiparametric latent factor model (SLFM) is an instance of LMC,
in which latent functions are defined by GPs~\cite{teh:semiparametric}.
Unfortunately, these multivariate models
cannot be straightforwardly used for
modeling the areal data we focus on,
because they assume that
the data samples are observed at location points;
namely they do not have an essential mechanism,
i.e., the spatial aggregation constraints,
for handling data that has been aggregated over regions.

The proposed model is an extension of SLFM.
To handle the multivariate areal data,
we newly introduce an observation model with
the spatial aggregation process for all areal data sets;
this is represented by the integral of the mixed GP
over each corresponding region, as in block kriging.
We also derive the posterior GP,
which enables us to obtain the multivariate function
from the observed areal data sets.
Furthermore, the sharing of key information (i.e., covariance function)
can be used for {\em transfer learning} across a wide variety of areal data sets;
this allows our model to robustly estimate the spatial correlations for areal data sets
and to support areal data sets from multiple domains.

Multi-task GP models have recently and independently been proposed
for addressing similar problems~\cite{hamelijnck:multi,yousefi:multi}.
Main differences of our work from them are as follows:
1) Explicit derivation of the posterior GP given multivariate areal data;
2) transfer learning across multiple domains;
3) extensive experiments on real-world data sets
defined on the two-dimensional input space.

\section{Proposed Model}
\label{sec:proposal}
We propose SAGP (Spatially Aggregated Gaussian Processes),
a probabilistic model for inferring
the multivariate function 
from areal data sets with various granularities.
We first consider a formulation in the case of a single domain,
then we mention an extension to the case of multiple domains.

{\bf Areal data.}
We start by describing the areal data this study focuses on.
For simplicity, let us consider the case of a single domain (e.g., a city).
Assume that we have a wide variety of areal data sets from the same domain
and each data set is associated with one of the geographical partitions
that have various granularities.
Let $S$ be the number of kinds of areal data sets.
Let ${\mathcal X} \subset {\mathbb R}^2$ denote an input space 
(e.g., a total region of a city), 
and ${\vect x} \in {\mathcal X}$ denote an input variable,
represented by its coordinates (e.g., latitude and longitude).
For $s = 1,\ldots,S$,
the partition ${\mathcal P}_s$ of ${\mathcal X}$ is
a collection of disjoint subsets,
called {\em regions}, of ${\mathcal X}$.
Let $|{\mathcal P}_s|$ be the number of regions in ${\mathcal P}_s$.
For $n = 1,\ldots,|{\mathcal P}_s|$,
let ${\mathcal R}_{s,n} \in {\mathcal P}_s$ denote the $n$-th region in ${\mathcal P}_s$.
Each areal observation is represented by the pair $({\mathcal R}_{s,n}, y_{s,n})$,
where $y_{s,n} \in {\mathbb R}$ is a value
associated with the $n$-th region ${\mathcal R}_{s,n}$.
Suppose that we have $S$ areal data sets
$\{ ({\mathcal R}_{s,n}, y_{s,n}) \mid s = 1,\ldots,S; n = 1,\ldots,|{\mathcal P}_s|  \}$.

{\bf Formulation for the case of a single domain.}
In the proposed model,
the functions for the respective areal data sets on the continuous space
are assumed to be the dependent Gaussian process (GP) with multivariate outputs.
We first construct the multivariate dependent GP by linearly mixing some independent latent GPs.
Consider $L$ independent GPs,
\vspace{-0.15\baselineskip}
\begin{flalign}
 g_l({\vect x})
 &\sim {\mathcal{GP}}
 \left(\nu_l({\vect x}), \gamma_l({\vect x}, {\vect x}^\prime) 
 \right),
 \quad l = 1,\ldots,L,
\end{flalign}
where $\nu_l({\vect x}) : {\mathcal X} \rightarrow {\mathbb R}$ and 
$\gamma_l({\vect x}, {\vect x}^\prime) : 
{\mathcal X} \times {\mathcal X} \rightarrow {\mathbb R}$
are a mean function and a covariance function, respectively,
for the $l$-th latent GP $g_l({\vect x})$, 
both of which are assumed integrable. 
Defining $f_s({\vect x})$ as the $s$-th GP,
the $S$-dimensional dependent GP
${\vect f}({\vect x}) = \left(f_1({\vect x}),\ldots,f_S({\vect x})\right)^\top$
is assumed to be modeled as a linear mixing of the $L$ independent latent GPs,
then ${\vect f}({\vect x})$ is given by
\vspace{-0.15\baselineskip}
\begin{flalign}
 {\vect f}({\vect x})
 &= {\bf W} {\vect g}({\vect x}) + {\vect n}({\vect x}),
\end{flalign}
where ${\vect g}({\vect x}) = \left( g_1({\vect x}),\ldots,g_L({\vect x}) \right)^\top$,
${\bf W}$ is an $S \times L$ weight matrix
whose $(s,l)$-entry $w_{s,l} \in {\mathbb R}$ is the weight of the $l$-th latent GP in the $s$-th data set,
and ${\vect n}({\vect x}) \sim {\mathcal{GP}}({\vect 0}, {\bf \Lambda}({\vect x}, {\vect x}^\prime))$ is
an $S$-dimensional zero-mean Gaussian noise process.
Here, ${\vect 0}$ is a column vector of 0's
and ${\bf \Lambda}({\vect x}, {\vect x}^\prime) 
= {\rm diag}(\lambda_1({\vect x}, {\vect x}^\prime),\ldots,\lambda_S({\vect x}, {\vect x}^\prime))$
with 
$\lambda_s({\vect x}, {\vect x}^\prime):{\mathcal X} \times {\mathcal X} \rightarrow {\mathbb R}$
being a covariance function for the $s$-th Gaussian noise process.
By integrating out ${\vect g}({\vect x})$, 
the multivariate GP ${\vect f}({\vect x})$ is given by
\vspace{-0.15\baselineskip}
\begin{flalign}
 {\vect f}({\vect x})
 &\sim {\mathcal{GP}}
 \left(
 {\vect m}({\vect x}), {\bf K}({\vect x}, {\vect x}^\prime)
 \right),
 \label{eq:f}
\end{flalign}
where the mean function ${\vect m}({\vect x}):{\mathcal X} \rightarrow {\mathbb R}^S$ 
is given by 
${\vect m}({\vect x}) = {\bf W} {\vect \nu}({\vect x})$.
The covariance function 
${\bf K}({\vect x}, {\vect x}^\prime): {\mathcal X} \times {\mathcal X} 
\rightarrow {\mathbb R}^{S \times S}$
is given by 
${\bf K}({\vect x}, {\vect x}^\prime)
= {\bf W} {\bf \Gamma}({\vect x}, {\vect x}^\prime) {\bf W}^\top + {\bf \Lambda}({\vect x}, {\vect x}^\prime)$.
Here, 
${\vect \nu}({\vect x}) = \left( \nu_1({\vect x}),\ldots,\nu_L({\vect x}) \right)^\top$
and
${\bf \Gamma}({\vect x}, {\vect x}^\prime) = 
{\rm diag} \left( \gamma_1({\vect x}, {\vect x}^\prime),\ldots,
\gamma_L({\vect x}, {\vect x}^\prime) \right)$.
The derivation of~(\ref{eq:f})
is described in Appendix~\ref{app:deriv_f} of Supplementary Material.
The $(s, s^\prime)$-entry of ${\bf K}({\vect x}, {\vect x}^\prime)$ is given by
\vspace{-0.25\baselineskip}
\begin{flalign}
k_{s,s^\prime}({\vect x}, {\vect x}^\prime) 
&= \delta_{s,s^\prime} \lambda_s({\vect x}, {\vect x}^\prime)
+ \sum_{l=1}^L w_{s,l} w_{s^\prime,l} \gamma_l({\vect x}, {\vect x}^\prime),
 \label{eq:covariance}
\end{flalign}
where $\delta_{\bullet, \bullet}$ represents Kronecker's delta;
$\delta_{Z, Z^\prime} = 1$ if $Z=Z^\prime$ and $\delta_{Z, Z^\prime} = 0$ otherwise.
The covariance function~(\ref{eq:covariance}) for the multivariate GP ${\vect f}({\vect x})$
is represented by the linear combination of the covariance functions
$\{\gamma_l({\vect x},{\vect x}^\prime)\}_{l=1}^L$ for the latent GPs.
The covariance functions for latent GPs are shared among all areal data sets,
which allows us to effectively learn the spatial correlation for each data set
by considering the dependences between data sets;
this is advantageous in the case where the number of observations is small,
that is, the spatial granularity of the areal data is coarse.
In this paper we focus on the case $L < S$, 
with the aim of reducing the number of free parameters
as this helps to avoid overfitting~\cite{teh:semiparametric}.
\begin{figure}[!tb]
 \begin{center}
  \subfigure[The case of a single domain.]
  {\includegraphics[height=25mm, bb=0 0 952 593]{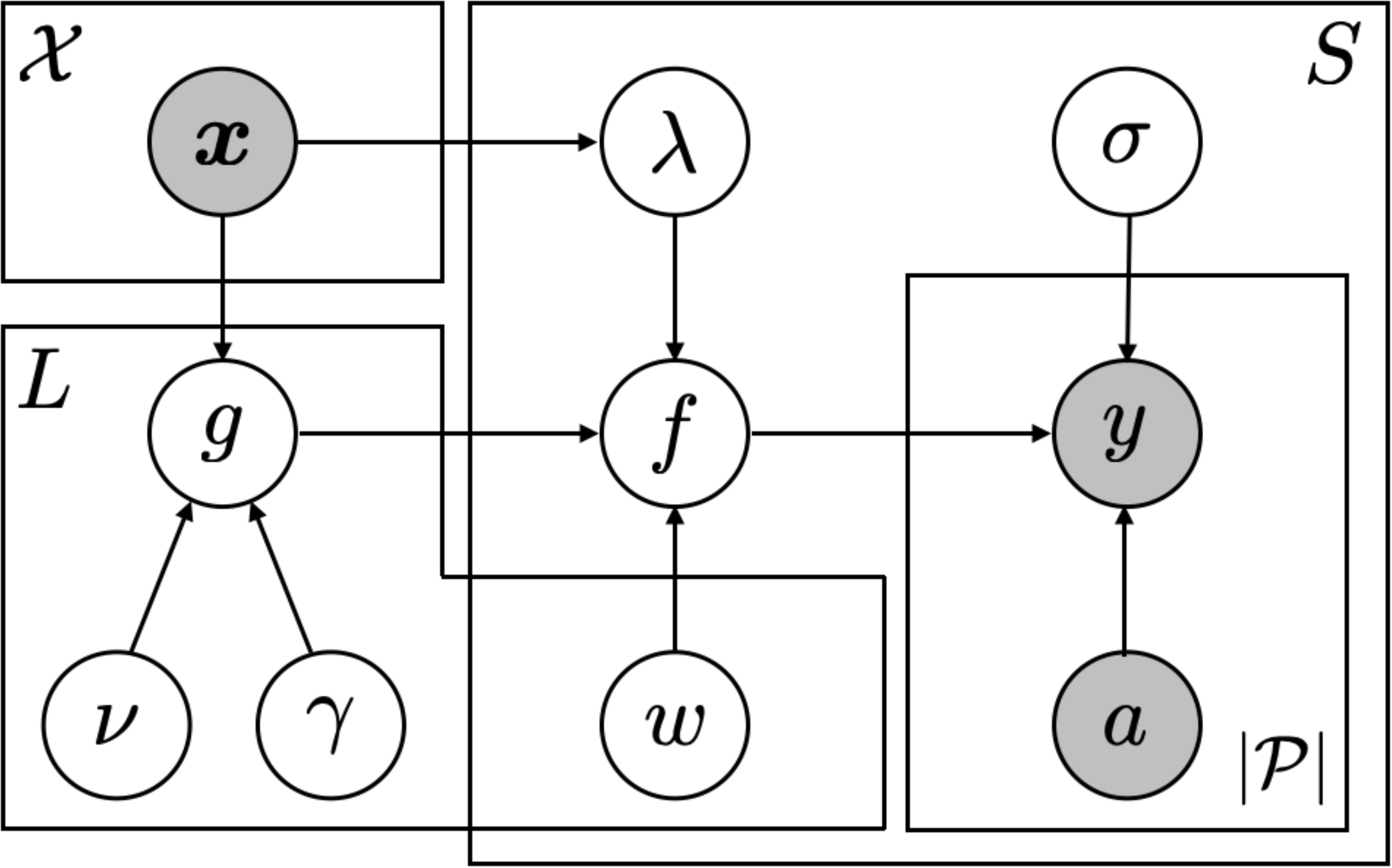}\label{fig:graphical_single}}
  \hspace{15pt}
  \subfigure[The case of two domains.]
  {\includegraphics[height=28mm, bb=0 0 1600 658]{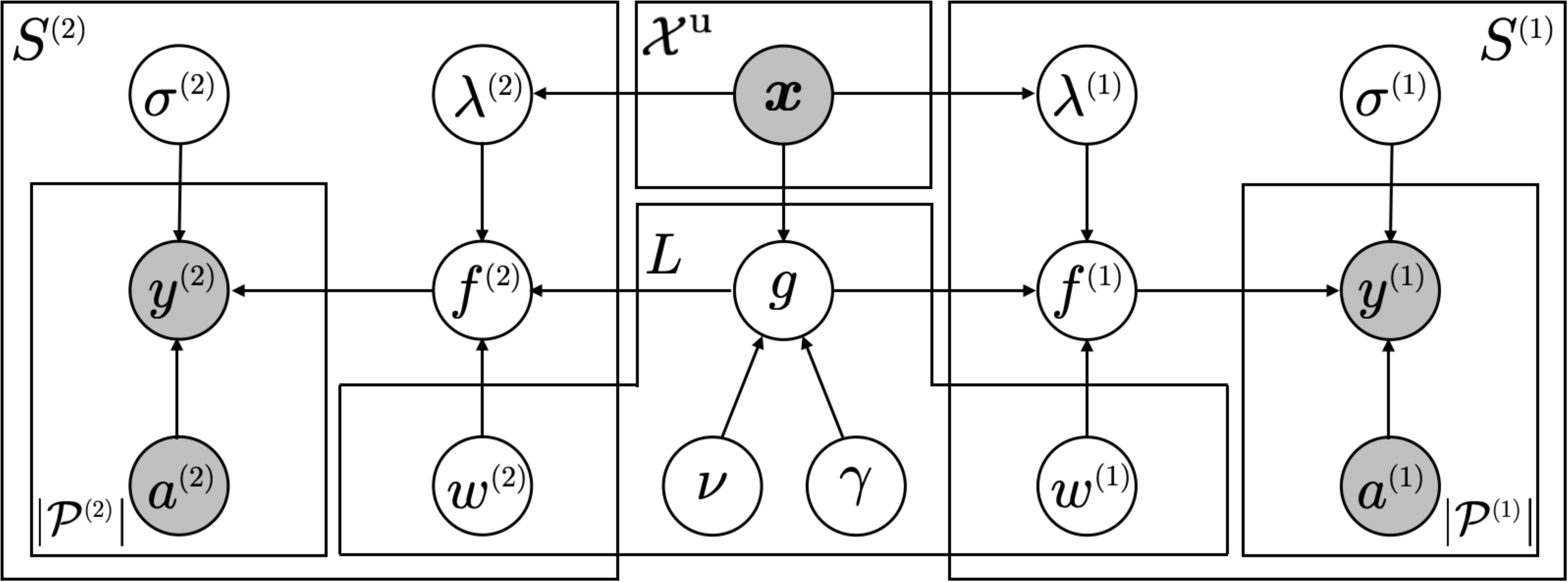}\label{fig:graphical_two}}
 \end{center}
\vspace{-10pt}
\caption{Graphical model representation of SAGP.}
\label{fig:graphical}
 \vspace{-5pt}
\end{figure}

The areal data are not associated with location points but with regions,
and their observations are obtained by spatially aggregating the original data.
To handle the multivariate areal data,
we design an observation model with a spatial aggregation process
for each of the areal data sets.
Let ${\vect y}_s =(y_{s,1},\ldots,y_{s,|{\mathcal P}_s|})$ be
a $|{\mathcal P}_s|$-dimensional vector consisting of the areal observations
for the $s$-th areal data set.
Let ${\vect y} = ({\vect y}_1, {\vect y}_2, \ldots, {\vect y}_S)^\top$ denote 
an $N$-dimensional vector consisting of the observations for all areal data sets,
where $N=\sum_{s=1}^S |{\mathcal P}_s|$ is the total number of areal observations.
Each areal observation is assumed to be obtained
by integrating the mixed GP ${\vect f}({\vect x})$ over the corresponding region;
${\vect y}$ is generated from the Gaussian distribution\footnote{%
  We here assume that the integral appearing in~\eqref{eq:generative_y}
  is well-defined.
  It should be noted that without additional assumptions
  sample paths of a Gaussian process are in general not integrable. 
  See Appendix~\ref{appendix:cont} of Supplementary Material
  for discussion on the conditions
  under which the integral is well-defined.},
\vspace{-0.25\baselineskip}
\begin{flalign}
 {\vect y} \mid {\vect f}({\vect x})
 &\sim {\mathcal N}
 \left(
 {\vect y} \Bigm| \int_{\mathcal X} {\bf A}({\vect x}){\vect f}({\vect x})\,d{\vect x}, 
 {\bf \Sigma}
 \right),
 \label{eq:generative_y}
\end{flalign}
where ${\bf A}({\vect x}) : {\mathcal X} \rightarrow {\mathbb R}^{N \times S}$ is represented by
\vspace{-0.25\baselineskip}
\begin{flalign}
 {\bf A}({\vect x})
 &= \left(
 \begin{array}{cccc}
  {\vect a}_{1}({\vect x}) & {\vect 0} & \cdots & {\vect 0} \\
  {\vect 0} & {\vect a}_{2}({\vect x}) & \cdots & {\vect 0} \\
  \vdots & \vdots & \ddots & \vdots \\
  {\vect 0} & {\vect 0} & \cdots & {\vect a}_S({\vect x})
 \end{array}
 \right),
\end{flalign}
in which ${\vect a}_s({\vect x}) = 
\left( a_{s,1}({\vect x}),\ldots,a_{s,|{\mathcal P}_s|}({\vect x}) \right)^\top$,
whose entry $a_{s,n}({\vect x})$ is a nonnegative weight function
for spatial aggregation over region ${\mathcal R}_{s,n}$.
This formulation does not depend on the particular choice of $\{a_{s,n}({\vect x})\}$, provided that they are integrable.
If one takes, for region ${\mathcal R}_{s,n}$, 
\vspace{-0.15\baselineskip}
\begin{flalign}
 a_{s,n}({\vect x})
 &=\frac{{\mathbbm 1}({\vect x} \in {\mathcal R}_{s,n})}
 {\int_{{\mathcal X}} {\mathbbm 1}({\vect x}^\prime \in {\mathcal R}_{s,n})\,d{\vect x}^\prime},
 \label{eq:average_model}
\end{flalign}
where ${\mathbbm 1} (\bullet)$ is the indicator function;
${\mathbbm 1}(Z) =1$ if $Z$ is true and ${\mathbbm 1}(Z) =0$ otherwise,
then $y_{s,n}$ is the average of $f_s({\vect x})$ over ${\mathcal R}_{s,n}$.
We may also consider other aggregation processes
to suit the property of the areal observations,
including simple summation and population-weighted averaging
over ${\mathcal R}_{s,n}$.
${\bf \Sigma} = {\rm diag}(\sigma_1^2 {\bf I},\ldots,\sigma_S^2 {\bf I})$ in (\ref{eq:generative_y}) is
an $N \times N$ block diagonal matrix,
where $\sigma_s^2$ is the noise variance for the $s$-th GP,
and ${\bf I}$ is the identity matrix.
Figure~\ref{fig:graphical_single} shows a graphical model representation of SAGP,
where shaded and unshaded nodes indicate observed and latent variables, respectively.

{\bf Extension to the case of multiple domains.}
It is possible to apply SAGP to areal data sets from multiple domains
by assuming that observations are conditionally independent
given the latent GPs $\{g_l({\vect x})\}_{l=1}^L$.
The graphical model representation of SAGP shown in Figure~\ref{fig:graphical_two} is
for the case of two domains.
The superscript in Figure~\ref{fig:graphical_two} is the domain index,
and ${\mathcal X}^{\rm u}$ is the union of the input spaces for both domains.
Although ${\vect y}^{(1)}$ and ${\vect y}^{(2)}$ in Figure~\ref{fig:graphical_two} 
are not directly correlated across domains,
the shared covariance functions
$\{\gamma_l({\vect x}, {\vect x}^\prime)\}_{l=1}^L$
for the latent GPs can be learned
by transfer learning based on the data sets from multiple domains;
thus the spatial correlation for each data set
could be more appropriately output
via the covariance functions,
even if we have only a few data sets available for a single domain.
SAGP can be extended to the case of more domains in a similar fashion.

\section{Inference}
\label{sec:inference}
Given the areal data sets,
we aim to derive the posterior GP on the basis of a Bayesian inference procedure.
The posterior GP can be used
for predicting data values at any location point in the continuous space.
The model parameters,
${\bf W}$, ${\bf \Lambda}({\vect x}, {\vect x}^\prime)$,
${\bf \Sigma}$, ${\vect \nu}({\vect x})$, ${\bf \Gamma}({\vect x}, {\vect x}^\prime)$,
are estimated by maximizing the marginal likelihood,
in which multivariate GP ${\vect f}({\vect x})$ is analytically integrated out;
we then construct the posterior GP by using the estimated parameters.

{\bf Marginal likelihood.}
Consider the case of a single domain.
Given the areal data ${\vect y}$,
the marginal likelihood is given by
\vspace{-0.15\baselineskip}
\begin{flalign}
 p({\vect y})
 &= {\mathcal N}
 \left(
 {\vect y} \mid {\vect \mu}, {\bf C}
 \right),
 \label{eq:marginal}
\end{flalign}
where we analytically integrate out the GP prior ${\vect f}({\vect x})$.
Here, ${\vect \mu}$ is an $N$-dimensional mean vector represented by
\vspace{-0.15\baselineskip}
\begin{flalign}
 {\vect \mu}
 &= \int_{\mathcal X} {\bf A}({\vect x}) {\vect m}({\vect x})\,d{\vect x},
 \label{eq:marginal_mean}
\end{flalign}
which is the integral of the mean function ${\vect m}({\vect x})$
over the respective regions for all areal data sets.
${\bf C}$ is an $N \times N$ covariance matrix represented by
\vspace{-0.15\baselineskip}
\begin{flalign}
 {\bf C}
 &= \iint_{{\mathcal X} \times {\mathcal X}}
 {\bf A}({\vect x}) {\bf K}({\vect x}, {\vect x}^\prime) {\bf A}({\vect x}^\prime)^\top
 \,d{\vect x}\,d{\vect x}^\prime + {\vect \Sigma}.
 \label{eq:marginal_cov}
\end{flalign}
It is an $S \times S$ block matrix
whose $(s,s^\prime)$-th block ${\bf C}_{s,s^\prime}$ is 
a $|{\mathcal P}_s| \times |{\mathcal P}_{s^\prime}|$ matrix represented by
\vspace{-0.15\baselineskip}
\begin{flalign}
 {\bf C}_{s,s^\prime}
 &=\iint_{{\mathcal X} \times {\mathcal X}} 
 k_{s,s^\prime}({\vect x},{\vect x}^\prime) 
 {\vect a}_s({\vect x}) 
 {\vect a}_{s^\prime}({\vect x}^\prime)^\top 
 \,d{\vect x}\,d{\vect x}^\prime 
 + \delta_{s,s^\prime} \sigma_s^2 {\bf I}.
 \label{eq:marginal_cov_s}
\end{flalign}
Equation~(\ref{eq:marginal_cov_s}) provides the region-to-region covariance
matrices in the form of the double integral
of the covariance function
$k_{s,s^\prime}(\bm{x},\bm{x}^\prime)$
over the respective pairs of regions in
$\mathcal{P}_s\times\mathcal{P}_{s'}$;
this conceptually corresponds to aggregation of the covariance function values
that are calculated at the infinite pairs of location points
in the corresponding regions.
Since the integrals over regions cannot be calculated analytically,
in practice we use a numerical approximation of these integrals.
Details are provided at the end of this section.
This formulation allows for accurately evaluating
the covariance between regions considering their shapes;
this is extremely helpful 
as some input data are likely to originate from irregularly shaped regions
(e.g., extremely elongated).
By maximizing the logarithm of the marginal likelihood~(\ref{eq:marginal}),
we can estimate the parameters of SAGP.

{\bf Transfer learning across multiple domains.}
Consider the case of $V$ domains.
Let $\{{\vect y}^{(v)}\}_{v=1}^V$ denote the collection of data sets for the $V$ domains.
In SAGP, the observations for different domains are assumed to be conditionally independent
given the shared latent GPs $\{g_l({\vect x})\}_{l=1}^L$;
the marginal likelihood for $V$ domains is thus given by the product of those for the $V$ domains:
\vspace{-0.15\baselineskip}
\begin{flalign}
 p\left( {\vect y}^{(1)}, {\vect y}^{(2)}, \ldots, {\vect y}^{(V)} \right)
 &= \prod_{v=1}^V {\mathcal N}
 \left(
 {\vect y}^{(v)} \bigm| {\vect \mu}^{(v)}, {\bf C}^{(v)}
 \right),
 \label{eq:marginal_multi}
\end{flalign}
where ${\vect \mu}^{(v)}$ and ${\bf C}^{(v)}$ are
the mean vector and the covariance matrix for the $v$-th domain, respectively.
Estimation of model parameters based on~(\ref{eq:marginal_multi})
allows for transfer learning 
across the areal data sets from multiple domains 
via the shared covariance functions. 

{\bf Posterior GP.}
We have only to consider the case of a single domain,
because the derivation of the posterior GP
can be conducted independently for each domain.
Given the areal data ${\vect y}$ and the estimated parameters,
the posterior GP ${\vect f}^\ast({\vect x})$ is given by
\vspace{-0.15\baselineskip}
\begin{flalign}
 {\vect f}^\ast({\vect x})
 &\sim {\mathcal{GP}}
 \left(
 {\vect m}^\ast({\vect x}), {\bf K}^\ast({\vect x}, {\vect x}^\prime)
 \right),
 \label{eq:posterior}
\end{flalign}
where ${\vect m}^\ast({\vect x}): {\mathcal X} \rightarrow {\mathbb R}^S$
and ${\bf K}^\ast({\vect x}, {\vect x}^\prime): {\mathcal X} \times {\mathcal X} 
\rightarrow {\mathbb R}^{S \times S}$ are
the mean function and the covariance function for ${\vect f}^\ast({\vect x})$, respectively.
Defining 
${\bf H}({\vect x}) : {\mathcal X} \rightarrow {\mathbb R}^{N \times S}$ as
\vspace{-0.15\baselineskip}
\begin{flalign}
 {\bf H}({\vect x})
 &= \int_{{\mathcal X}} 
 {\bf A}({\vect x}^\prime) {\bf K}({\vect x}^\prime, {\vect x})\,d{\vect x}^\prime,
 \label{eq:H}
\end{flalign}
which consists of the point-to-region covariances,
which are the covariances between any location point $\bm{x}$ and
the respective regions in all areal data sets,
the mean function ${\vect m}^\ast({\vect x})$
and the covariance function ${\bf K}^\ast({\vect x}, {\vect x}^\prime)$ are given by
\vspace{-0.15\baselineskip}
\begin{flalign}
 {\vect m}^\ast({\vect x})
 &= {\vect m}({\vect x}) +  
 {\bf H}({\vect x})^\top {\bf C}^{-1} ({\vect y} - {\vect \mu}),
 \label{eq:post_mean}\\
 {\bf K}^\ast({\vect x}, {\vect x}^\prime)
 &= {\bf K}({\vect x}, {\vect x}^\prime) - 
 {\bf H}({\vect x})^\top {\bf C}^{-1} {\bf H}({\vect x}^\prime),
 \label{eq:post_var}
\end{flalign}
respectively. 
We can predict the data value at any location point
by using the mean function~(\ref{eq:post_mean}).
The second term in~(\ref{eq:post_mean}) shows that the predictions
are calculated by considering the spatial correlations
and the dependences between areal data sets, simultaneously.
By using the covariance function~(\ref{eq:post_var}),
we can also evaluate the prediction uncertainty.
Derivation of the posterior GP is detailed
in Appendix~\ref{appendix:posteriorGP} of Supplementary Material. 

{\bf Approximation of the integral over regions.}
The integrals over regions in~(\ref{eq:marginal_mean}), (\ref{eq:marginal_cov_s}), and (\ref{eq:H})
cannot be performed analytically;
thus we approximate these integrals by using
sufficiently fine-grained square grid cells.
We divide input space ${\mathcal X}$ into square grid cells,
and take ${\mathcal G}_{s,n}$ to be the set of grid points
that are contained in region ${\mathcal R}_{s,n}$.
Let us consider the approximation of the integral in the covariance matrix~(\ref{eq:marginal_cov_s}).
The $(n,n^\prime)$-entry $C_{s,s^\prime}(n,n^\prime)$
of ${\bf C}_{s,s^\prime}$ is approximated as follows:
\vspace{-0.15\baselineskip}
\begin{flalign}
 {\bf C}_{s,s^\prime}(n,n^\prime)
 &=\iint_{{\mathcal X} \times {\mathcal X}} k_{s,s^\prime}({\vect x}, {\vect x}^\prime) 
 a_{s,n}({\vect x}) a_{s^\prime,n^\prime}({\vect x}^\prime)\,d{\vect x}\,d{\vect x}^\prime 
 + \delta_{s,s^\prime} \delta_{n,n^\prime} \sigma_s \\
 &\approx \frac{1}{|{\mathcal G}_{s,n}|} \frac{1}{|{\mathcal G}_{s^\prime,n^\prime}|}
 \sum_{i \in {\mathcal G}_{s,n}} \sum_{j \in {\mathcal G}_{s^\prime,n^\prime}}
 k_{s,s^\prime}(i,j) + \delta_{s,s^\prime} \delta_{n,n^\prime} \sigma_s,
 \label{eq:approx}
\end{flalign}
where we use the formulation of the region-average-observation model~(\ref{eq:average_model}).
The integrals in~(\ref{eq:marginal_mean}) and (\ref{eq:H}) can be approximated
in a similar way.
Letting $|{\mathcal G}|$ denote the number of all grid points,
the computational complexity of ${\bf C}_{s,s^\prime}$~(\ref{eq:marginal_cov_s}) is $O(|{\mathcal G}|^2)$;
assuming the constant weight $a_{s,n}({\vect x}) = a_{s,n}$ (e.g., region average),
the computational complexity can be reduced to $O(|{\mathcal P}_s| |{\mathcal P}_{s^\prime}| |{\mathcal D|})$,
where $|{\mathcal D}|$ is 
the cardinality of the set of distinct distance values between grid points.
Here, we use the property that $k_{s,s^\prime}(i,j)$ in~(\ref{eq:approx})
depends only on the distance between $i$ and $j$.
This is useful for reducing the computation time and the memory requirement.
The average computation times for inference were
1728.2 and 115.1 seconds for the data sets
from New York City and Chicago, respectively;
the experiments were conducted
on a 3.1 GHz Intel Core i7.

\section{Experiments}
{\bf Data.}
We evaluated SAGP using 10 and 3 real-world areal data sets
from two cities, New York City and Chicago, respectively.
They were obtained from NYC Open Data~\footnote{https://opendata.cityofnewyork.us}
and Chicago Data Portal~\footnote{https://data.cityofchicago.org/}.
We used a variety of areal data sets
including poverty rate, air pollution rate, and crime rate.
Each data set is associated with one of the predefined geographical partitions with various granularities:
UHF42 (42), community district (59), police precinct (77), and zip code (186) in New York City;
police precinct (25) and community district (77) in Chicago,
where each number in parentheses denotes the number of regions in the corresponding partition.
In the experiments, the data were normalized
so that each variable in each city has
zero mean and unit variance.
Details about the real-world data sets are provided in Appendix~\ref{app:data} 
of Supplementary Material. 

{\bf Refinement task.}
We examined the task of refining coarse-grained areal data
by using multiple areal data sets with various granularities.
To evaluate the performance in predicting the fine-grained areal data,
we first picked up one target data set and used its coarser version
for learning model parameters;
then we predicted the original fine-grained target data by using the learned model.
Note that the fine-grained target data was used only for evaluating the refinement performance;
we did not use them in the inference process.
The target data sets were
poverty rate (5, 59), PM2.5 (5, 42), crime (5, 77) in New York City
and poverty rate (9, 77) in Chicago,
where each pair of numbers in parentheses denotes
the numbers of regions in the coarse- and the fine-grained partitions, respectively.
Defining $s^\prime$ as the index of the target data set,
the evaluation metric is the mean absolute percentage error (MAPE)
of the fine-grained target values,
$\frac{1}{|{\mathcal P}_{s^\prime}|} \sum_{n \in {\mathcal P}_{s^\prime}}
\left| (y^{\rm true}_{s^\prime,n} - y^\ast_{s^\prime,n}) / {y^{\rm true}_{s^\prime,n}} \right|$,
where $y^{\rm true}_{s^\prime,n}$ is the true value associated with the $n$-th region
in the target fine-grained partition;
$y^\ast_{s^\prime,n}$ is its predicted value, obtained
by integrating the $s^\prime$-th function $f^\ast_{s^\prime}({\vect x})$
of the posterior GP ${\vect f}^\ast({\vect x})$~(\ref{eq:posterior})
over the corresponding target fine-grained region.

{\bf Setup of the proposed model.}
In our experiments, we used zero-mean Gaussian processes
as the latent GPs $\{g_l(\bm{x})\}_{l=1}^L$, i.e.,
$\nu_l({\vect x}) = 0$ for $l = 1,\ldots,L$.
We used the following squared-exponential kernel as the covariance function for the latent GPs,
$\gamma_l({\vect x}, {\vect x}^\prime)
= \alpha^2_l \exp \left(
- \|{\vect x} - {\vect x}^\prime\|^2 / 2\beta^2_l
\right)$,
where $\alpha^2_l$ is a signal variance
that controls the magnitude of the covariance,
$\beta_l$ is a scale parameter
that determines the degrees of spatial correlation,
and $\|\cdot\|$ is the Euclidean norm.
Here, we set $\alpha^2_l = 1$ because the variance can already be
represented by scaling the columns of ${\bf W}$.
For simplicity, the covariance function for the Gaussian noise process
${\vect n}({\vect x}, {\vect x}^\prime)$ is set to
${\bf \Lambda}({\vect x}, {\vect x}^\prime) = {\rm diag}(\lambda_1^2 \delta({\vect x}-{\vect x}^\prime),\ldots,\lambda_S^2 \delta({\vect x}-{\vect x}^\prime))$,
where $\delta(\bullet)$ is Dirac's delta function.
The model parameters, ${\bf W}$, $\{\lambda_s\}_{s=1}^S$, ${\bf \Sigma}$, $\{\beta_l\}_{l=1}^L$,
were learned by maximizing the logarithm of
the marginal likelihood (\ref{eq:marginal}) or (\ref{eq:marginal_multi}) 
using the L-BFGS method~\cite{liu:on} implemented in SciPy (\url{https://www.scipy.org/}).
For approximating the integral over regions (see (\ref{eq:approx})),
we divided a total region of each city into sufficiently fine-grained square grid cells,
the size of which was 300~m $\times$ 300~m for both cities;
the resulting sets of grid points ${\mathcal G}$
for New York City and Chicago
consisted of 9,352 and 7,400 grid points, respectively.
The number $L$ of the latent GPs was chosen from $\{1,\ldots,S\}$
via leave-one-out cross-validation~\cite{bishop:pattern};
the validation error was obtained using each held-out coarse-grained data value.
Here, the validation was conducted on the basis of the coarse-grained target areal data;
namely we did not use the fine-grained target data in the validation process.

{\bf Baselines.}
We compared the proposed model, SAGP, with naive Gaussian process regression (GPR)~\cite{carl:gaussian},
two-stage GP-based model (2-stage GP)~\cite{tanaka:refining},
and semiparametric latent factor model (SLFM)~\cite{teh:semiparametric}.
GPR predicts the fine-grained target data simply 
from just the coarse-grained target data.
2-stage GP is one of the latest regression-based models.
SLFM is the multivariate GP model;
SAGP is regarded as the extension of SLFM.
GPR and SLFM assume that data samples are observed at location points.
We thus associate each areal observation with the centroid of the region.
This simplification is also used for modeling the auxiliary data sets in~\cite{tanaka:refining}.

\begin{table}[!t]
 \small
 \caption{MAPE and standard errors for the prediction
 of fine-grained areal data (a single city).
 The numbers in parentheses denote the number $L$ of the latent GPs
 determined by the validation procedure.
 The single star ($\star$) and the double star ($\star \star$)
 indicate significant difference between SAGP and other models at the levels of $P$ values of $< 0.05$ and $< 0.01$,
 respectively.}
 \vspace{-10pt}
 \label{tb:MAPE}
 \begin{center}
  \begin{tabular}{l c c c c c} \toprule
   &\multicolumn{3}{c}{New York City} &\multicolumn{1}{c}{Chicago} \\
   \cmidrule(l){2-4} \cmidrule(l){5-5}
   &Poverty rate &PM2.5 &Crime &Poverty rate \\ \midrule
   GPR& 0.344 $\pm$ 0.046 (--) & 0.072 $\pm$ 0.010 (--) & 0.860 $\pm$ 0.102 (--) & 0.599 $\pm$ 0.099 (--) \\[1pt]
   2-stage GP& 0.210 $\pm$ 0.022 (--) & 0.042 $\pm$ 0.005 (--) & 0.454 $\pm$ 0.075 (--) & 0.380 $\pm$ 0.060 (--) \\[1pt]
   SLFM& 0.207 $\pm$ 0.025 (4) & 0.036 $\pm$ 0.005 (6) & 0.401 $\pm$ 0.053 (2) & 0.335 $\pm$ 0.052 (2) \\[1pt]
   SAGP& {\bf 0.177} $\pm$ {\bf 0.019}$^{\star \star}$ (3) & {\bf 0.030} $\pm$ {\bf 0.005}$^{\star}$ (5) & {\bf 0.379} $\pm$ {\bf 0.055}$^{\star \star}$ (3) & {\bf 0.278} $\pm$ {\bf 0.032}$^{\star \star}$ (2) \\[1pt]
   \bottomrule
  \end{tabular}
 \end{center}
\vspace{-11pt}
\end{table}
\begin{figure}[!t]
 \begin{minipage}{.2\textwidth}
  \subfigure[SAGP]
  {\includegraphics[width=24mm, bb=0 0 1000 1000]{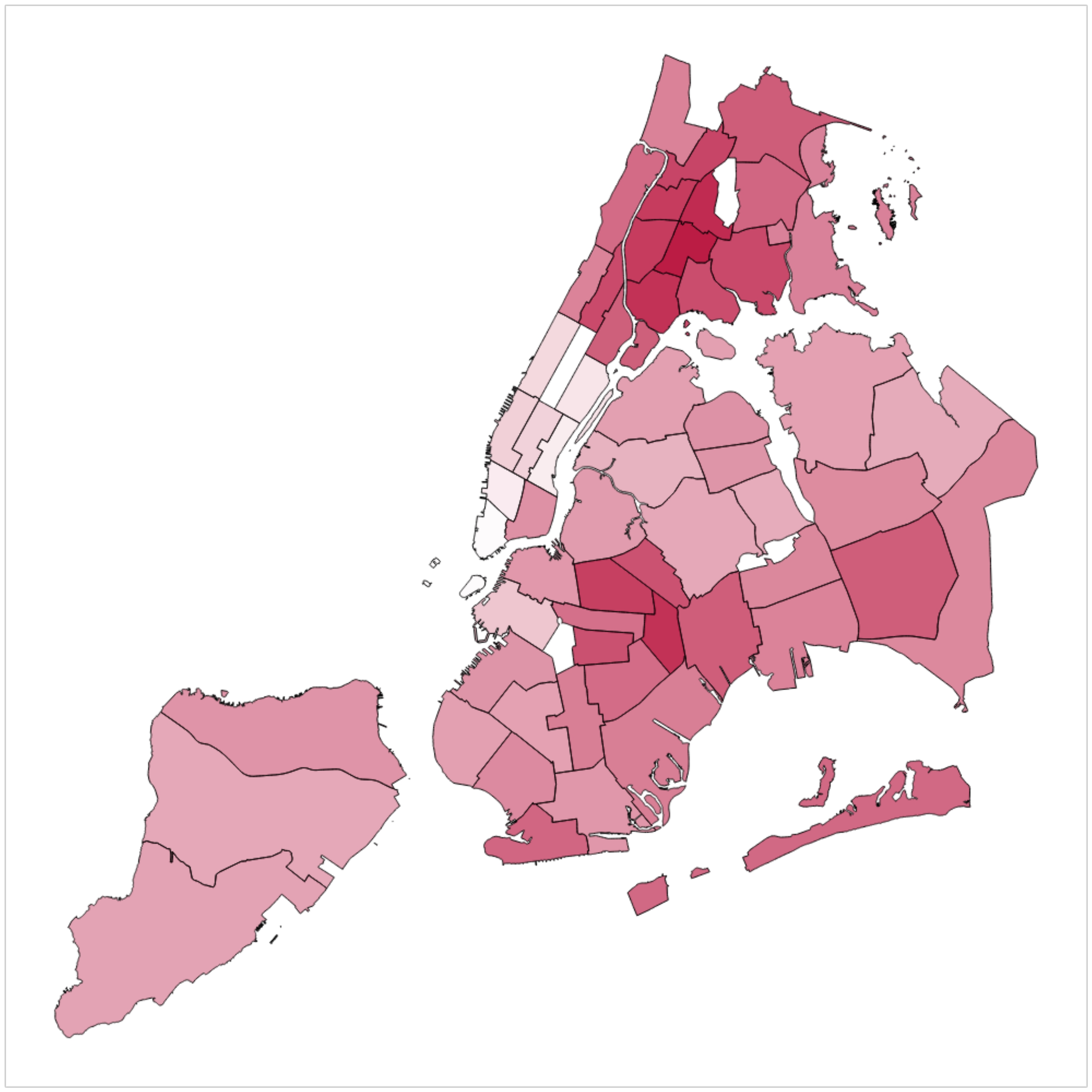}
  \label{fig:sagp_poverty}}
  \subfigure[SLFM]
  {\includegraphics[width=24mm, bb=0 0 1000 1000]{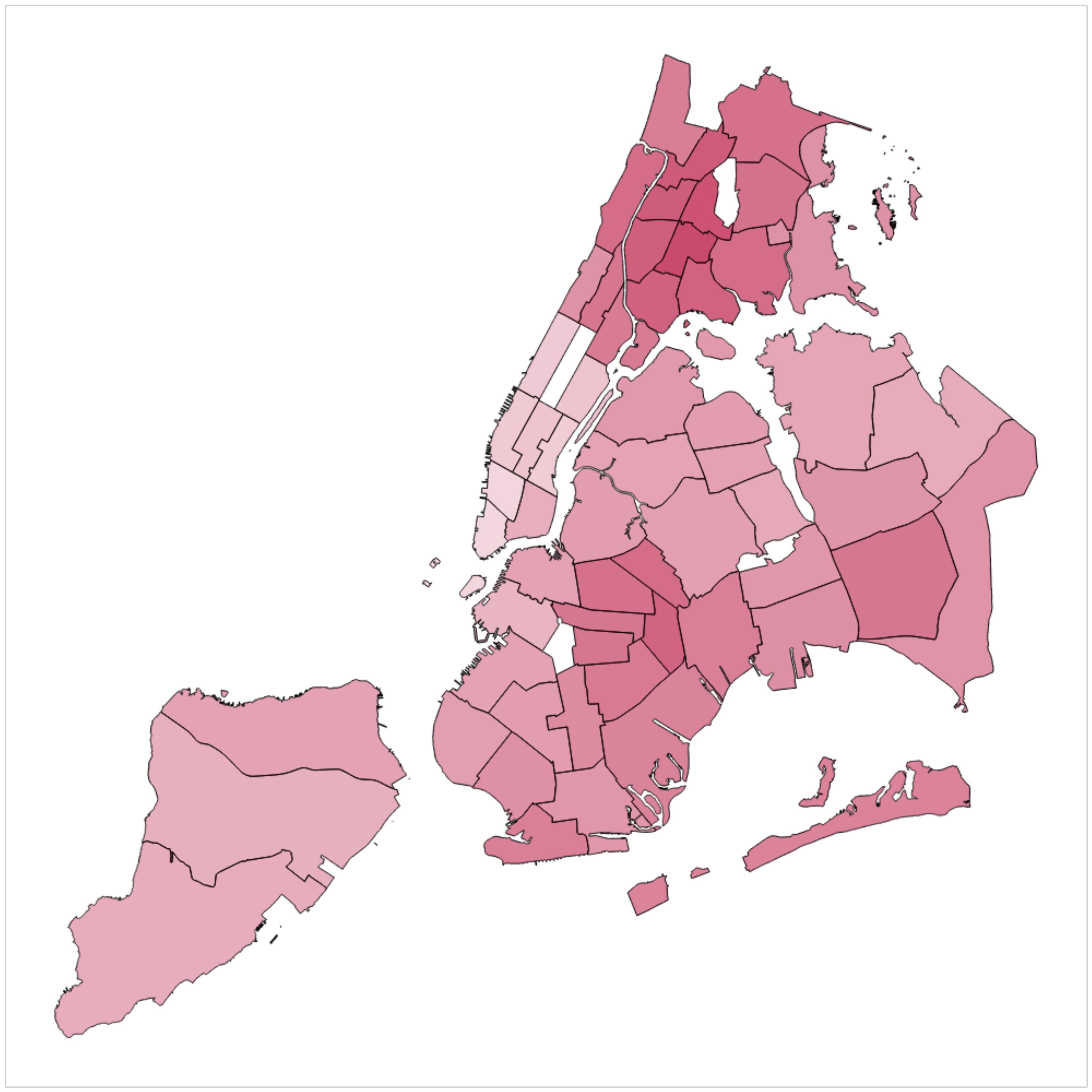}
  \label{fig:slfm_poverty}}
 \end{minipage}
 \hfill
 \hspace{10pt}
 \begin{minipage}{.8\textwidth}
  \subfigure[Visualization of the estimated parameters ${\bf W}$ and $\{\beta_l\}_{l=1}^L$.]
  {\includegraphics[width=98mm, bb=0 0 1800 1050]{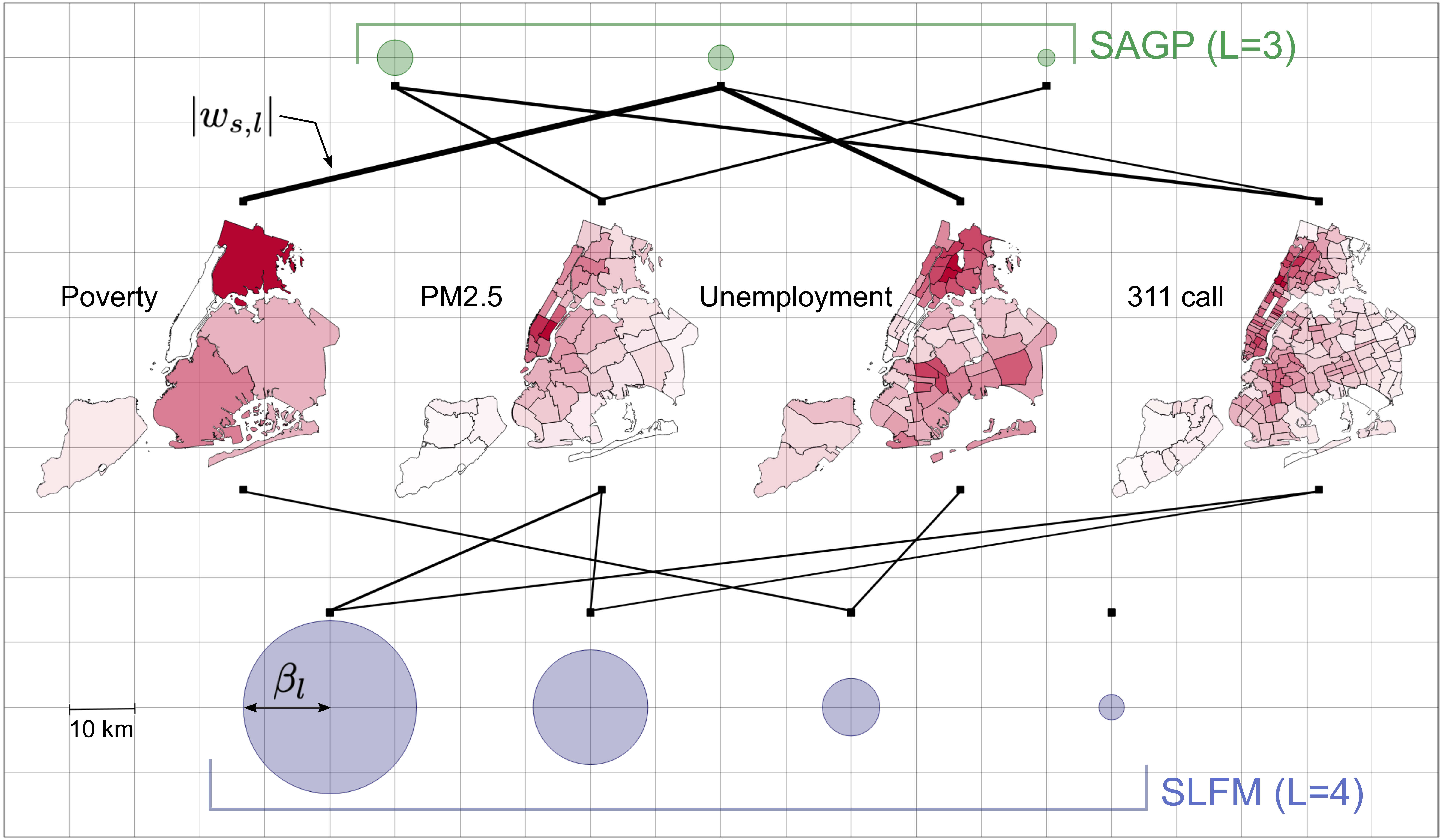}
  \label{fig:vis_param}}
 \end{minipage}
 \vspace{-11pt}
 \caption{(a,b) Refined poverty rate data in NYC,
 and (c) Visualization of the estimated parameters
 when predicting the poverty rate data in NYC.
 The radii of green and blue circles equal the values of $\beta_l$
 estimated by SAGP and SLFM, respectively.
 The edge widths are proportional to the absolute weights $|w_{s,l}|$
 estimated by the respective models.
 Here, we omitted those edges whose absolute weights were lower than a threshold.}
\vspace{-14pt}
\end{figure}
{\bf Results for the case of a single city.}
Table~\ref{tb:MAPE} shows MAPE and standard errors for GPR, 2-stage GP, SLFM, and SAGP.
For all data sets, SAGP achieved better performance in
refining coarse-grained areal data;
the differences between SAGP and the baselines were statistically
significant (Student's t-test).
These results show that SAGP can utilize the areal data sets
with various granularities from the same city
to accurately predict the refined data.
The results for all data sets from both cities
are shown in Appendix~\ref{app:results} of Supplementary Material.

Figures~\ref{fig:sagp_poverty} and~\ref{fig:slfm_poverty} show the refinement results of
SAGP and SLFM for the poverty rate data in New York City.
Here, the predictive values of each model were normalized
to the range $[0,1]$, and darker hues represent regions with higher values.
Compared with the true data in Figure~\ref{fig:areal},
SAGP yielded more accurate fine-grained data than SLFM.
Figure~\ref{fig:vis_param} visualizes the mixing weights
${\bf W}$ and the scale parameters $\{\beta_l\}_{l=1}^L$
estimated by SAGP and SLFM when predicting the fine-grained poverty rate data in New York City,
where we picked up 4 areal data sets: Poverty rate, PM2.5, unemployment rate, and the number of 311 calls;
their observations were also shown.
One observes that the scale parameters estimated by SAGP are relatively small
compared with those estimated by SLFM,
presumably because
the spatial aggregation process 
incorporated in SAGP 
effectively separates intrinsic spatial correlations
and apparent smoothing effects due to the spatial aggregation
to yield areal observations. 
A comparison of the estimated weights in Figure~\ref{fig:vis_param} shows that
SAGP emphasized the useful dependences between data sets,
e.g., the strong correlation between the poverty rate data and the unemployment rate data.

\begin{wrapfigure}{r}{68mm}
 \vspace{-5pt}
 \begin{tabular}{c}
  \begin{minipage}{\linewidth}
   {\includegraphics[width=30mm]{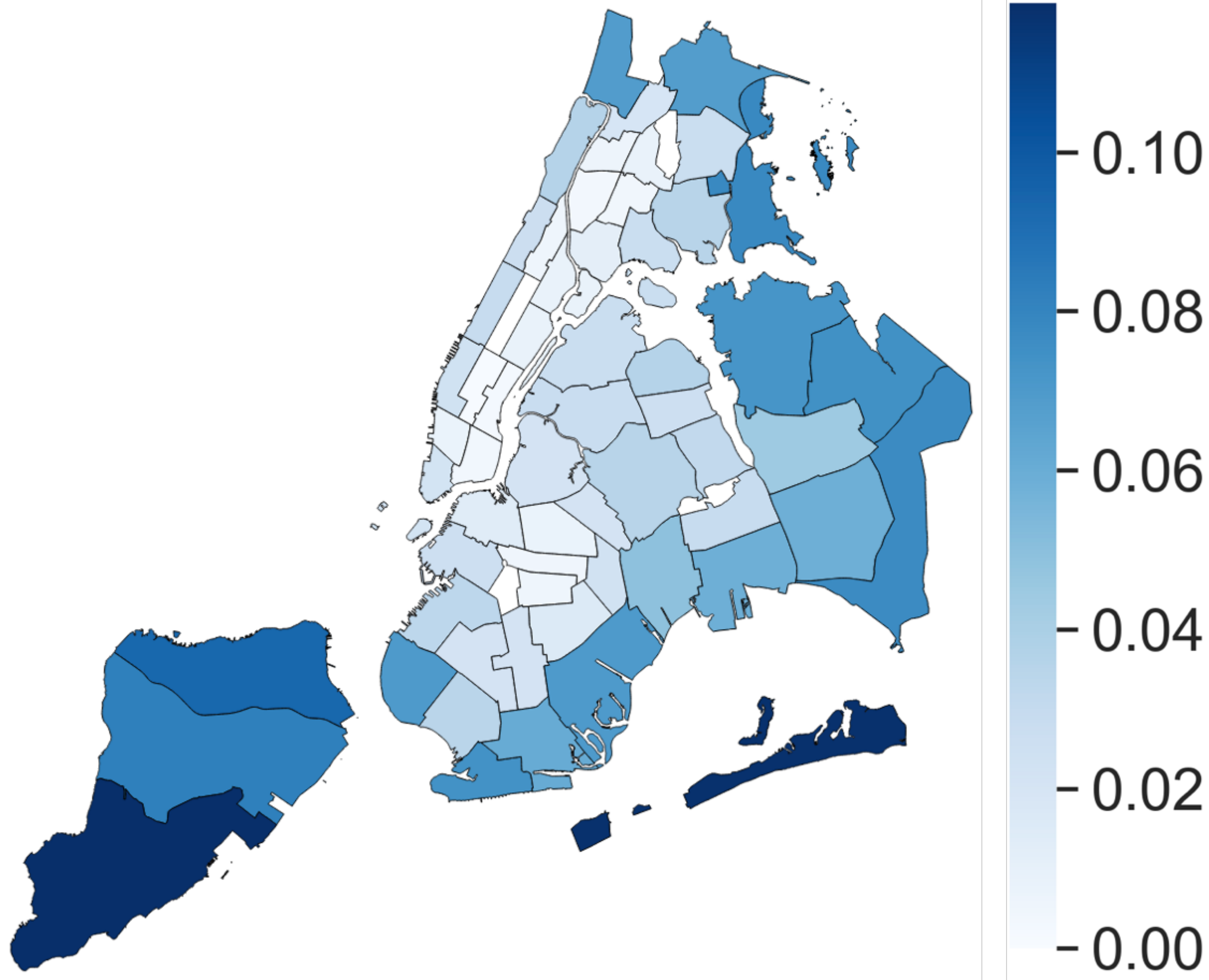}\label{fig:}}
   \hspace{5pt}
   {\includegraphics[width=30mm]{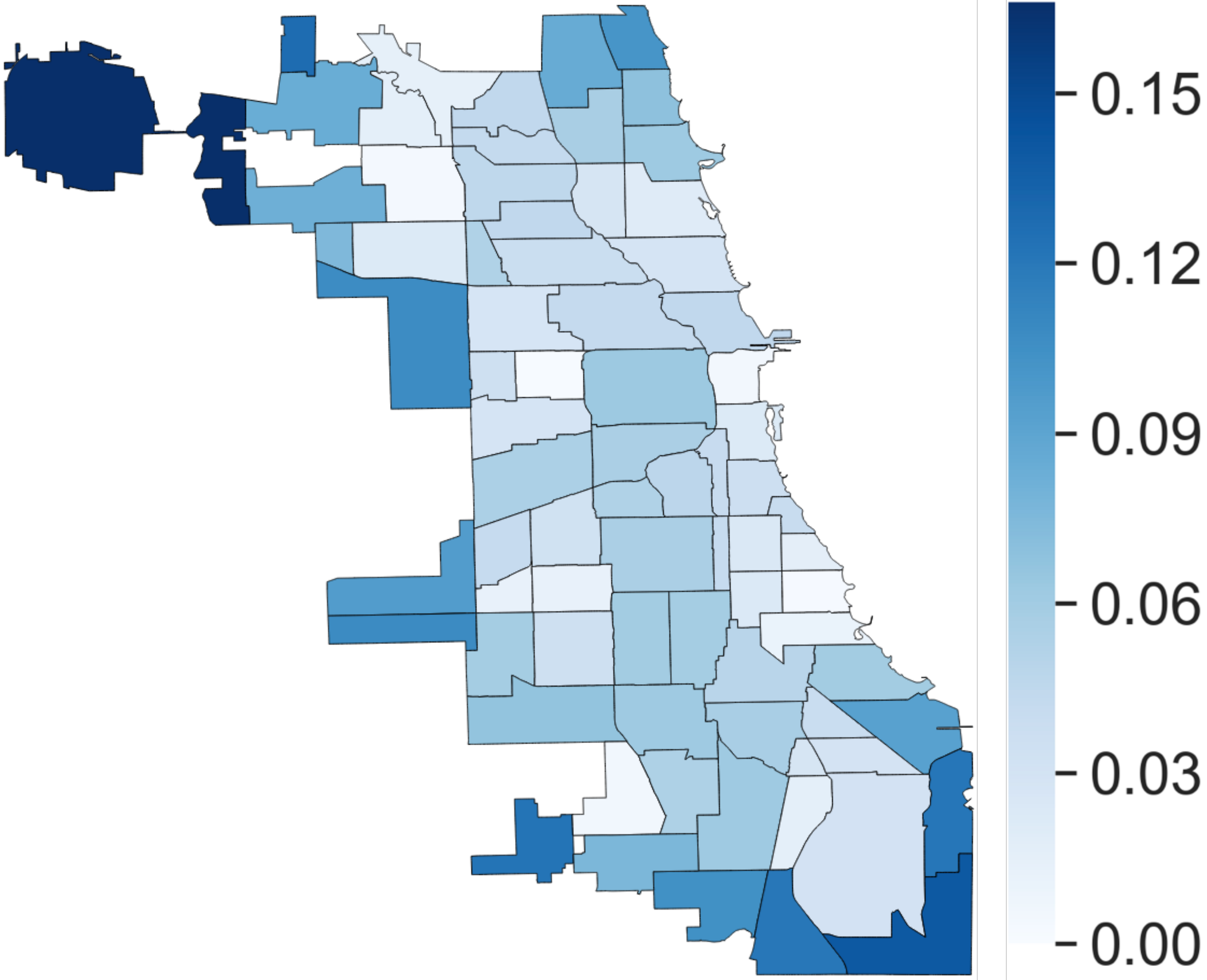}\label{fig:}}
  \end{minipage} 
 \end{tabular}
 \caption{Variance of the posterior GP with SAGP
 for predicting the poverty rate in New York City (Left)
 and Chicago (Right), respectively.}
 \label{fig:variance}
\end{wrapfigure}
One benefit of SAGP is that all predictions
associated with the target regions have
uncertainty estimates,
where the prediction variance can be calculated by
integrating the covariance function
${\bf K}^\ast(\bm{x}, \bm{x}^\prime)$~(\ref{eq:post_var})
of the posterior GP~(\ref{eq:posterior})
over the corresponding target region.
Figure~\ref{fig:variance} visualizes the variance with SAGP
in the prediction of the poverty rate 
in New York City and Chicago, respectively.
One observes that the variances
at the regions located at the edge of the city
tend to have larger values
compared with those inside the city.
This is reasonable because
extrapolation is generally more difficult than interpolation. 
These uncertainty estimates are useful
in that the predictions may help guide
policy and planning in a city
even if validation of them is difficult.

\begin{figure}[!t]
 \vspace{+3pt}
 \footnotesize
  \def\@captype{table}
  \begin{minipage}[l]{.35\textwidth}
   \tblcaption{MAPE and standard errors for the prediction of the fine-grained data (two cities).}
   \vspace{+2pt}
  \begin{tabular}{l c} \toprule
   &\multicolumn{1}{c}{Chicago} \\ \cmidrule(l){2-2}
   &Poverty rate \\ \midrule
   SLFM (trans) & 0.328 $\pm$ 0.050 (6) \\[1pt]
   SAGP (trans) & {\bf 0.219} $\pm$ {\bf 0.023} (4) \\[1pt]
   \bottomrule
  \end{tabular}
  \label{tb:MAPE_trans}
  \end{minipage}
 \hfill
 \hspace{10pt}
 \begin{minipage}[r]{.65\textwidth}
  \subfigure[True]
  {\includegraphics[width=25mm, bb=0 0 1000 1000]{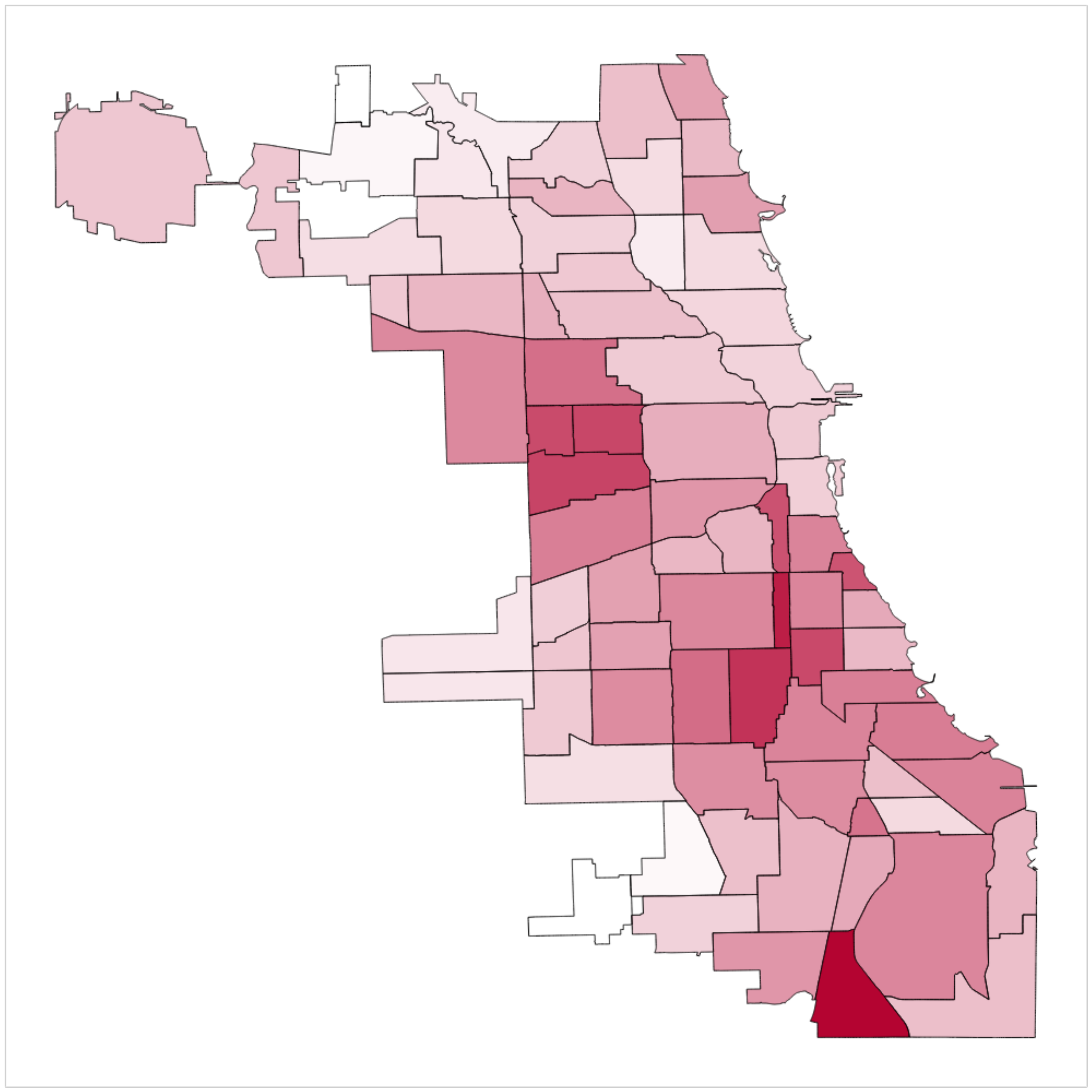}\label{fig:}}
  \hspace{5pt}
  \subfigure[SAGP (trans)]
  {\includegraphics[width=25mm, bb=0 0 1000 1000]{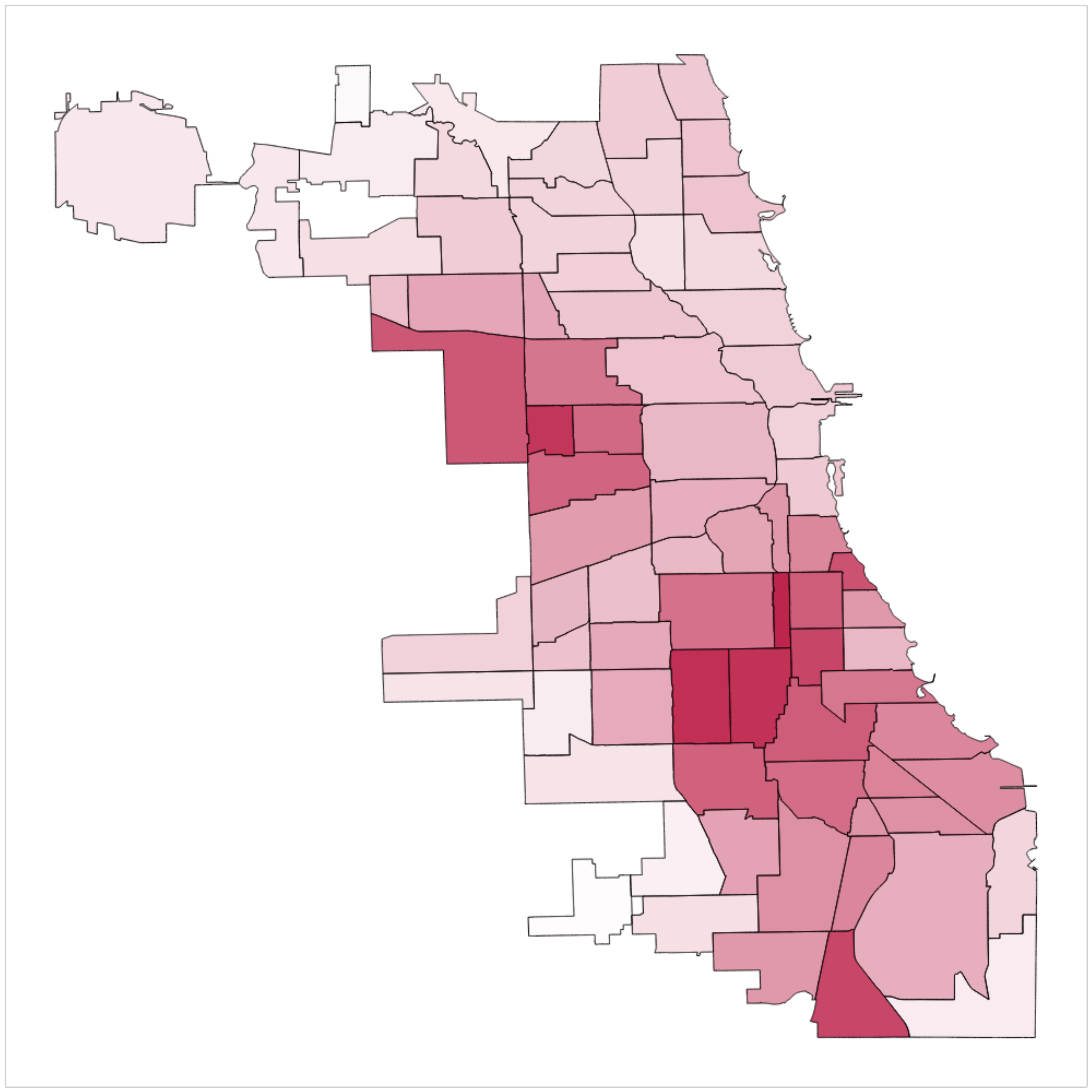}\label{fig:}}
  \hspace{5pt}
  \subfigure[SLFM (trans)]
  {\includegraphics[width=25mm, bb=0 0 1000 1000]{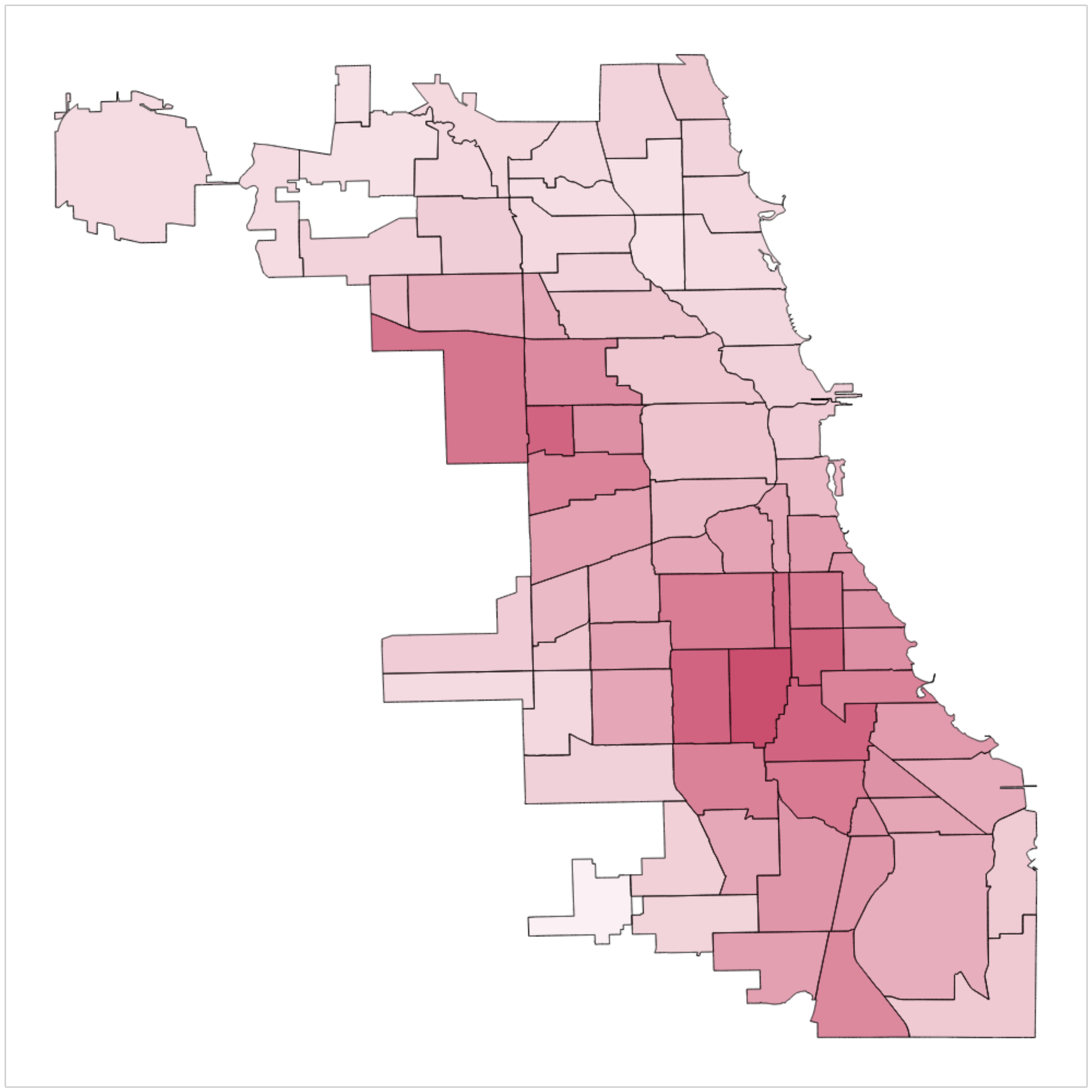}\label{fig:}}
  \vspace{-7pt}
  \caption{Refined poverty rate data in Chicago.}
  \label{fig:comparison}
 \end{minipage}
 \vspace{-14pt}
\end{figure}

{\bf Results for the case of two cities.}
SLFM and SAGP can be used for transfer learning across multiple cities,
which is more advantageous in such a situation that
we have only a few data sets available on a single city.
We here show the results of refining the poverty rate data in Chicago
with simultaneously utilizing the data sets from New York City.
Table~\ref{tb:MAPE_trans} shows MAPE and standard errors for SLFM (trans) and SAGP (trans).
Comparing Tables~\ref{tb:MAPE} and~\ref{tb:MAPE_trans},
one observes that SAGP (trans) attained improved refinement performance compared with SLFM (trans)
and models trained with only the data in a single city (i.e., Chicago).
the differences between SAGP (trans) and the other models were statistically
significant (Student's t-test, $P$ value of $< 0.01$).
This result shows that SAGP (trans) transferred knowledge across the cities,
and yielded better refinement results
even if there are only a few data sets available on the target city.
Figure~\ref{fig:comparison} shows the refinement results for
the poverty rate data in Chicago.
We illustrate the true data on the left in Figure~\ref{fig:comparison},
and the predictions attained by SAGP (trans) and SLFM (trans) on the right.
As shown, SAGP (trans) better identified the key regions compared with SLFM (trans).

\section{Conclusion}
This paper has proposed the Spatially Aggregated Gaussian Processes
for inferring the multivariate function 
from multiple areal data sets with various granularities.
To handle multivariate areal data,
we design an observation model with the spatial aggregation process
for each areal data set,
which is the integral of the Gaussian process over the corresponding region.
We have confirmed that our model can accurately refine the coarse-grained areal data,
and improve the refinement performance by using the areal data sets from multiple cities.

There are several avenues that can be explored in future work.
First, we can introduce nonlinear link functions,
as in warped GP~\cite{edward:warped}, and/or alternative likelihoods;
this might help handle some kinds of observations (e.g., rates).
Second, we can use scalable variational inference with inducing points,
similar to~\cite{titsias:variational},
for large-scale data sets.
Finally, our formulation provides a general framework
for modeling aggregated data
and offers a potential research direction;
for instance, it has the ability
to consider data aggregated
on a higher dimensional input space,
e.g., spatio-temporal aggregated data.

\bibliographystyle{plain}

\newpage
\appendix
\begin{center}
  \large\textbf{Supplementary Material: Spatially Aggregated Gaussian Processes with Multivariate Areal Outputs}
\end{center}
\section{Derivation of the multivariate GP $\vect{f}(\vect{x})$}
\label{app:deriv_f}
In this appendix, we show that the process $\bm{f}(\bm{x})$
defined via~(2) is itself a multivariate GP
with mean function $\bm{m}(\bm{x})=\mathbf{W}\bm{\nu}(\bm{x})$
and covariance function $\mathbf{K}(\bm{x},\bm{x}')
=\mathbf{W\Gamma}(\bm{x},\bm{x}')\mathbf{W}^\top+\mathbf{\Lambda}(\bm{x},\bm{x}')$.
To prove that $\bm{f}(\bm{x})$ is indeed a multivariate GP,
one has only to show that, 
for an arbitrary $k\in\{1,2,\ldots\}$
and an arbitrary set of $k$ points $\bm{x}_1,\ldots,\bm{x}_k\in\mathcal{X}$, 
$\bar{\bm{f}}=(\bm{f}(\bm{x}_1),\ldots,\bm{f}(\bm{x}_k))^\top$
is a multivariate Gaussian random variable. 
By the definition (2) of $\bm{f}(\bm{x})$, one has 
\begin{equation}
\label{eq:bar_f}
\bar{\bm{f}}=(\mathbf{I}\otimes\mathbf{W})\bar{\bm{g}}+\bar{\bm{n}},
\end{equation} 
where we let $\bar{\bm{g}}=(\bm{g}(\bm{x}_1),\ldots,\bm{g}(\bm{x}_k))^\top$ 
and $\bar{\bm{n}}=(\bm{n}(\bm{x}_1),\ldots,\bm{n}(\bm{x}_k))^\top$,
and where $\otimes$ denotes the Kronecker product. 
By the definition of Gaussian processes,
since $\bm{g}(\bm{x})$ and $\bm{n}(\bm{x})$ are Gaussian processes, 
$\bar{\bm{g}}$ and $\bar{\bm{n}}$ are multivariate Gaussian random variables.
Since~\eqref{eq:bar_f} shows that $\bar{\bm{f}}$ is a linear combination of
the multivariate Gaussian random variables $\bar{\bm{g}}$ and $\bar{\bm{n}}$,
it is itself multivariate Gaussian,
irrespective of the choice of $\bm{x}_1,\ldots,\bm{x}_k$.
This in turn shows that $\bm{f}(\bm{x})$ is again a multivariate Gaussian process.

Mean of $\bm{f}(\bm{x}_i)$ is given by 
\begin{equation}
\mathbb{E}(\bm{f}(\bm{x}_i))=\mathbf{W}\mathbb{E}(\bm{g}(\bm{x}_i))
=\mathbf{W}\bm{\nu}(\bm{x}_i).
\end{equation}
Covariance of $\bm{f}(\bm{x}_i)$ and $\bm{f}(\bm{x}_j)$ is given by
\begin{align}
\mathrm{Cov}(\bm{f}(\bm{x}_i),\bm{f}(\bm{x}_j))
&=\mathbb{E}((\bm{f}(\bm{x}_i)-\mathbf{W}\bm{\nu}(\bm{x}_i))
(\bm{f}(\bm{x}_j)-\mathbf{W}\bm{\nu}(\bm{x}_j))^\top)
\nonumber\\
&=\mathbb{E}((\mathbf{W}(\bm{g}(\bm{x}_i)-\bm{\nu}(\bm{x}_i))+\bm{n}(\bm{x}_i))(\mathbf{W}(\bm{g}(\bm{x}_j)-\bm{\nu}(\bm{x}_j))+\bm{n}(\bm{x}_j))^\top)
\nonumber\\
&=\mathbf{W\Gamma}(\bm{x}_i,\bm{x}_j)\mathbf{W}^\top+\mathbf{\Lambda}(\bm{x}_i,\bm{x}_j).
\end{align}
These show that the mean function $\bm{m}(\bm{x})$
and the covariance function $\mathbf{K}(\bm{x},\bm{x}')$ of
the multivariate Gaussian process $\bm{f}(\bm{x})$ are given by
$\bm{m}(\bm{x})=\mathbf{W}\bm{\nu}(\bm{x})$
and $\mathbf{K}(\bm{x},\bm{x}')=\mathbf{W\Gamma}(\bm{x},\bm{x}')
\mathbf{W}^\top+\mathbf{\Lambda}(\bm{x},\bm{x}')$, respectively. 

\section{Derivation of the posterior GP $\vect{f}^\ast(\vect{x})$}
\label{appendix:posteriorGP}
In this appendix, we derive the posterior Gaussian process $\bm{f}^\ast(\bm{x})$
shown in Section 4.
We here assume 
that the integral appearing in the definition
of the observation model~\eqref{eq:generative_y} is well-defined, 
and defer discussion on conditions for its well-definedness
to Appendix~\ref{appendix:cont}. 
Let $\bm{f}(\bm{x})\sim\mathcal{GP}(\bm{m}(\bm{x}),\mathbf{K}(\bm{x},\bm{x}'))$ be a multivariate GP defined on $\mathcal{X}\subset\mathbb{R}^2$
taking values in $\mathbb{R}^S$. 
For an arbitrary $k,k'\in\mathbb{N}$
and an arbitrary set of $(k+k')$ points
$\bm{x}_1,\ldots,\bm{x}_k,\bm{x}_1',\ldots,\bm{x}_{k'}'\in\mathcal{X}$,
let
\begin{equation}
  \hat{\bm{f}}=\left(\begin{array}{c}\bar{\bm{f}}\\\bar{\bm{f}}'
  \end{array}\right),
  \quad
  \bar{\bm{f}}=\left(\begin{array}{c}\bm{f}(\bm{x}_1)\\\vdots\\\bm{f}(\bm{x}_k)
  \end{array}\right),
  \quad
  \bar{\bm{f}}'=\left(\begin{array}{c}\bm{f}(\bm{x}_1')\\\vdots\\\bm{f}(\bm{x}_{k'}')
  \end{array}\right).
\end{equation}
By the definition of GP,
$\hat{\bm{f}}$ is a $(k+k')S$-dimensional Gaussian vector.
Let
\begin{equation}
  \hat{\bm{m}}=\left(\begin{array}{c}\bar{\bm{m}}\\\bar{\bm{m}}'
  \end{array}\right),
  \quad
  \bar{\bm{m}}=\left(\begin{array}{c}\bm{m}(\bm{x}_1)\\\vdots\\\bm{m}(\bm{x}_k)
  \end{array}\right),
  \quad
  \bar{\bm{m}}'=\left(\begin{array}{c}\bm{m}(\bm{x}_1')\\\vdots\\\bm{m}(\bm{x}_{k'}')
  \end{array}\right)
\end{equation}
and
\begin{equation}
  \hat{\mathbf{K}}=\left(\begin{array}{cc}
    \bar{\mathbf{K}} & \bar{\mathbf{K}}'^\top\\
    \bar{\mathbf{K}}' & \bar{\mathbf{K}}''
  \end{array}\right),
  \quad
  (\bar{\mathbf{K}})_{ij}=\mathbf{K}(\bm{x}_i,\bm{x}_j),
  \quad
  (\bar{\mathbf{K}}')_{ij}=\mathbf{K}(\bm{x}_i',\bm{x}_j),
  \quad
  (\bar{\mathbf{K}}'')_{ij}=\mathbf{K}(\bm{x}_i',\bm{x}_j')
\end{equation}
be the mean vector and the covariance matrix of $\hat{\bm{f}}$.
In the following,
we specifically assume that $\bm{x}_1',\ldots,\bm{x}_{k'}'$ are taken to be
grid points of a regular grid covering $\mathcal{X}$
and with the grid cell volume $\Delta$,
and consider Riemann sums to approximate those integrals on $\mathcal{X}$
appearing in the formulation of SAGP. 
We then take the limit $\Delta\to0$ to derive the posterior GP
given areal observations on 
$\bm{f}(\bm{x})\sim\mathcal{GP}(\bm{m}(\bm{x}),\mathbf{K}(\bm{x},\bm{x}'))$.

Consider the observation process yielding observations $\bm{y}$,
defined by
\begin{equation}
  \bm{y}=\bar{\mathbf{A}}\bar{\bm{f}}'+\bm{w},
\end{equation}
where
\begin{equation}
  \bar{\mathbf{A}}=\left(\begin{array}{ccc}
    \mathbf{A}(\bm{x}_1')&\cdots&\mathbf{A}(\bm{x}_{k'}')
  \end{array}\right)\Delta,
\end{equation}
and where $\bm{w}$ is an $S$-dimensional Gaussian noise vector
with mean zero and covariance $\mathbf{\Sigma}$.
One has
\begin{equation}
  \bar{\bm{\mu}}=\mathbb{E}(\bm{y})
  =\bar{\mathbf{A}}\bar{\bm{m}}'
\end{equation}
and
\begin{equation}
  \bar{\mathbf{C}}=\mathrm{Cov}(\bm{y})
  =\bar{\mathbf{A}}\bar{\mathbf{K}}''\bar{\mathbf{A}}^\top+\mathbf{\Sigma},
\end{equation}
respectively.
The posterior of $\bar{\bm{f}}$ given $\bm{y}$ is
known to be a multivariate Gaussian with mean
\begin{equation}
  \bar{\bm{m}}^*
  =\bar{\bm{m}}+\bar{\mathbf{H}}^\top\mathbf{C}^{-1}(\bm{y}-\bm{\mu})
\end{equation}
and covariance
\begin{equation}
  \bar{\mathbf{K}}^*
  =\bar{\mathbf{K}}-\bar{\mathbf{H}}^\top\mathbf{C}^{-1}\bar{\mathbf{H}},
\end{equation}
respectively, where $\bar{\mathbf{H}}=\bar{\mathbf{A}}\bar{\mathbf{K}}'$.

By regarding sums over the $k'$ terms as Riemann sums approximating
the corresponding integrals over $\mathcal{X}$,
in the limit $\Delta\to0$,
one can replace those sums over $k'$ terms
with the corresponding integrals over $\mathcal{X}$.
Specifically, one has 
\begin{align}
  \bm{y}
  &=\sum_{i=1}^{k'}\mathbf{A}(\bm{x}_i')\bm{f}(\bm{x}_i')\Delta
  +\bm{w}
  \nonumber\\
  &\to\int_{\mathcal{X}}\mathbf{A}(\bm{x})\bm{f}(\bm{x})\,d\bm{x}
  +\bm{w},
  \\
  \bar{\bm{\mu}}
  &=\sum_{i=1}^{k'}\mathbf{A}(\bm{x}_i')\bm{m}(\bm{x}_i')\Delta
  \nonumber\\
  &\to\int_{\mathcal{X}}\mathbf{A}(\bm{x})\bm{m}(\bm{x})\,d\bm{x}
  =\bm{\mu},
  \\
  \bar{\mathbf{C}}
  &=\sum_{i,j=1}^{k'}
  \mathbf{A}(\bm{x}_i')\mathbf{K}(\bm{x}_i',\bm{x}_j')
  \mathbf{A}(\bm{x}_j')^\top\Delta^2+\mathbf{\Sigma}
  \nonumber\\
  &\to\iint_{\mathcal{X}\times\mathcal{X}}
  \mathbf{A}(\bm{x})\mathbf{K}(\bm{x},\bm{x}')
  \mathbf{A}(\bm{x}')^\top\,d\bm{x}\,d\bm{x}'+\mathbf{\Sigma}
  =\mathbf{C},
\end{align}
showing that the mean vector $\bar{\bm{\mu}}$
and the covariance matrix $\bar{\mathbf{C}}$ of $\bm{y}$ are reduced
in this limit to the vector $\bm{\mu}$ and 
the matrix $\mathbf{C}$ defined in~\eqref{eq:marginal_mean}
and~(\ref{eq:marginal_cov}), respectively.
One also has
\begin{align}
  \bar{\mathbf{H}}
  &=\left(\begin{array}{ccc}
    \sum_{i=1}^{k'}\mathbf{A}(\bm{x}_i')\mathbf{K}(\bm{x}_i',\bm{x}_1)\Delta
    &\cdots&
    \sum_{i=1}^{k'}\mathbf{A}(\bm{x}_i')\mathbf{K}(\bm{x}_i',\bm{x}_k)\Delta
  \end{array}\right)
  \nonumber\\
  &\to\left(\begin{array}{ccc}
    \int_{\mathcal{X}}\mathbf{A}(\bm{x}')\mathbf{K}(\bm{x}',\bm{x}_1)\,d\bm{x}'
    &\cdots&
    \int_{\mathcal{X}}\mathbf{A}(\bm{x}')\mathbf{K}(\bm{x}',\bm{x}_k)\,d\bm{x}'
  \end{array}\right)
  \nonumber\\
  &=\left(\begin{array}{ccc}
    \mathbf{H}(\bm{x}_1)
    &\cdots&
    \mathbf{H}(\bm{x}_k)
  \end{array}\right),
\end{align}
where $\mathbf{H}(\bm{x})$ is defined in~\eqref{eq:H}.
The above calculation shows that in the limit $\Delta\to0$
the posterior process $\bm{f}^*(\bm{x})$ is a multivariate GP
with mean function $\bm{m}^*(\bm{x})$
and covariance function $\mathbf{K}^*(\bm{x},\bm{x}')$ given by
\eqref{eq:post_mean} and \eqref{eq:post_var}, respectively.

\section{On integrability}
\label{appendix:cont}
In this appendix, we discuss conditions for the observation
model~\eqref{eq:generative_y} to be well defined. 
Assume that $\mathcal{X}\subset\mathbb{R}^2$ is a bounded
Jordan-measurable set,
and that elements $a_{s,n}(\bm{x})$ of $\mathbf{A}(\bm{x})$
are Riemann integrable on $\mathcal{X}$. 
(The latter assumption is satisfied when $a_{s,n}(\bm{x})$
is defined as in~\eqref{eq:average_model}
with Jordan-measurable regions $\mathcal{R}_{s,n}$.) 
Since it is known that a continuous function on $\mathcal{X}$
is Riemann integrable on $\mathcal{X}$,
and that a product of Riemann integrable functions is again
Riemann integrable,
a sufficient condition for the observation model~\eqref{eq:generative_y} 
to be well defined is that the prior process
$\bm{f}(\bm{x})\sim\mathcal{GP}(\bm{m}(\bm{x}),\mathbf{K}(\bm{x},\bm{x}'))$
is sample-path continuous.

The assumption made in Section~\ref{sec:proposal}
of the integrability of $\nu_l(\bm{x})$, $l=1,\ldots,L$,
assures integrability of the mean function
$\bm{m}(\bm{x})=\mathbf{W}\bm{\nu}(\bm{x})$, 
which allows us to reduce integrability of the prior process
$\bm{f}(\bm{x})\sim\mathcal{GP}(\bm{m}(\bm{x}),\mathbf{K}(\bm{x},\bm{x}'))$
to sample-path continuity of the zero-mean process 
$\bm{f}(\bm{x})\sim\mathcal{GP}(\mathbf{0},\mathbf{K}(\bm{x},\bm{x}'))$ 
on $\mathcal{X}$.
A sufficient condition~\cite[Theorem 1.4.1]{adlertaylor2007}
for the sample-path continuity of the zero-mean Gaussian process 
is that for some $0<C<\infty$ and $\alpha,\eta>0$
\begin{equation}
k_{s,s}(\bm{x},\bm{x})+k_{s,s}(\bm{x}',\bm{x}')-2k_{s,s}(\bm{x},\bm{x}')
\le\frac{C}{|\log\|\bm{x}-\bm{x}'\||^{1+\alpha}}
\end{equation}
holds for all $s\in\{1,\ldots,S\}$
and for all $\bm{x},\bm{x}'$ with $\|\bm{x}-\bm{x}'\|<\eta$. 
If one uses the squared-exponential kernels for $\{\gamma_l\}$,
then one can confirm that the above condition is satisfied,
and consequently the observation model~\eqref{eq:generative_y}
is well defined. 

It should be noted that the sample-path continuity discussed above is
different from the mean-square (MS) continuity.
A process $\bm{f}(\bm{x})$ is said to be MS continuous at $\bm{x}=\bm{x}_*$
if for any sequence $\bm{x}_k$ converging to $\bm{x}_*$ as $k\to\infty$
it holds that $\mathbb{E}[\|\bm{f}(\bm{x}_k)-\bm{f}(\bm{x}_*)\|^2]\to0$
as $k\to\infty$.
A necessary and sufficient condition for a random field to be
MS continuous at $\bm{x}_*$ is that
its covariance function $\mathbf{K}(\bm{x},\bm{x}')$ is continuous
at the point $\bm{x}=\bm{x}'=\bm{x}_*$~\cite[Appendix 10A]{papoulis1991},
which in the case of Gaussian processes
is weaker than the above sufficient condition for the sample-path
continuity. 

\section{Description of real-world areal data sets}
\label{app:data}
We used the real-world areal data sets from
NYC Open Data~\footnote{https://opendata.cityofnewyork.us}
and Chicago Data Portal~\footnote{https://data.cityofchicago.org/}
to evaluate the proposed model.
These data sets are collected and released
for improving city environments,
and consist of a variety of categories including
social indicators, land use, and air quality.
Details of the areal data sets we used in the experiments
are listed in Table~\ref{tb:data}.
The number of data sets in New York City and Chicago are 10 and 3, respectively.
Each data set is associated with one of the predefined geographical partitions.
The number of partition types in New York City and Chicago are 4 and 2, respectively.
Table~\ref{tb:data} shows
the respective partition names and the number of regions in the corresponding partition.
These data sets are gathered once a year
at the time ranges shown in Table~\ref{tb:data};
the values of data were divided by the number of observation times.
Then, the data were normalized
so that each variable in each city has
zero mean and unit variance.
\begin{table}[tb]
 \caption{Real-world areal data sets.}
 \label{tb:data}
 \begin{center}
  \subtable[New York City]{\label{tb:}
  \begin{tabular}{l l r r} \toprule
   Data &Partition &\#regions &Time range \\ \midrule
   PM2.5 &UHF42 &42 &2009 -- 2010 \\
   Poverty rate &Community district &59 &2009 -- 2013 \\
   Unemployment rate &Community district &59 &2009 -- 2013 \\
   Mean commute &Community district &59 &2009 -- 2013 \\
   Population &Community district &59 &2009 -- 2013 \\
   Recycle diversion rate &Community district &59 &2009 -- 2013 \\
   Crime &Police precinct &77 &2010 -- 2016 \\
   Fire incident &Zip code &186 &2010 -- 2016 \\
   311 call &Zip code &186 &2010 -- 2016 \\
   Public telephone &Zip code &186 &2016 \\
   \bottomrule
  \end{tabular}
  }
  \\
  \subtable[Chicago]{\label{tb:}
  \begin{tabular}{l l r r} \toprule
   Data &Partition &\#regions &Time range \\ \midrule
   Crime &Police Precinct &25 &2012 \\
   Poverty rate &Community district &77 &2008 -- 2012 \\
   Unemployment rate &Community district &77 &2008 -- 2012 \\
   \bottomrule
  \end{tabular}
  }
 \end{center}
\end{table}

\section{Results}
\label{app:results}
Table~\ref{tb:result_nyc} shows MAPE and standard errors
for GPR, 2-stage GP, SLFM, and SAGP,
where the experiments for Crime data set in Chicago have not been conducted
because the coarser version for training is not available online.
For all data sets, 
SAGP achieved the comparable or better performance than the other methods.

\begin{table}[!t]
 \small
 \caption{MAPE and standard errors for the prediction of fine-grained areal data
 in New York City and Chicago. The numbers in parentheses denote the number $L$ of the latent GPs
 estimated by the validation procedure.
 The single star ($\star$) and the double star ($\star \star$)
 indicate significant difference between SAGP and other models at the levels of $P$ values of $< 0.05$ and $< 0.01$,
 respectively.}
 \vspace{-10pt}
 \label{tb:result_nyc}
 \begin{center}
 \subtable[New York City]{\label{tb:}
  \begin{tabular}{l c c c c c} \toprule
   &GPR &2-stage GP &SLFM &SAGP \\ \midrule
   PM2.5 & 0.072 $\pm$ 0.010 (--) & 0.042 $\pm$ 0.005 (--) & 0.036 $\pm$ 0.005 (6) & {\bf 0.030 $\pm$ 0.005}$^{\star}$ (5) \\[1pt]
   Poverty rate & 0.344 $\pm$ 0.046 (--) & 0.210 $\pm$ 0.022 (--) & 0.207 $\pm$ 0.025 (4) & {\bf 0.177 $\pm$ 0.019}$^{\star\star}$ (3) \\[1pt]
   Unemployment rate & 0.319 $\pm$ 0.036 (--) & 0.193 $\pm$ 0.021 (--) & 0.195 $\pm$ 0.024 (3) & {\bf 0.165 $\pm$ 0.020}$^{\star}$  (3) \\[1pt]
   Mean commute & 0.131 $\pm$ 0.020 (--) & 0.068 $\pm$ 0.009 (--) & 0.057 $\pm$ 0.007 (4) & {\bf 0.050 $\pm$ 0.007} (6) \\[1pt]
   Population & 0.577 $\pm$ 0.104 (--) & 0.389 $\pm$ 0.033 (--) & 0.337 $\pm$ 0.039 (3) & {\bf 0.295 $\pm$ 0.033}$^{\star}$ (3) \\[1pt]
   Recycle diversion rate & 0.353 $\pm$ 0.049 (--) & 0.236 $\pm$ 0.034 (--) & 0.222 $\pm$ 0.032 (4) &{\bf 0.211 $\pm$ 0.029} (4) \\[1pt]
   Crime & 0.860 $\pm$ 0.102 (--) & 0.454 $\pm$ 0.075 (--) & 0.401 $\pm$ 0.053 (2) & {\bf 0.379 $\pm$ 0.055}$^{\star \star}$ (3) \\[1pt]
   Fire incident & 1.097 $\pm$ 0.097 (--) & 0.746 $\pm$ 0.084 (--) & 0.500 $\pm$ 0.052 (4) & {\bf 0.396 $\pm$ 0.038}$^{\star\star}$ (3) \\[1pt]
   311 call & 0.083 $\pm$ 0.004 (--) & 0.070 $\pm$ 0.004 (--) & 0.061 $\pm$ 0.004 (6) & {\bf 0.052 $\pm$ 0.003}$^{\star\star}$ (3) \\[1pt]
   Public telephone & 0.131 $\pm$ 0.008 (--) & 0.083 $\pm$ 0.008 (--) & 0.086 $\pm$ 0.008 (4) & {\bf 0.080 $\pm$ 0.007} (6) \\[1pt]
   \bottomrule
  \end{tabular}}
  \\
 \subtable[Chicago]{\label{tb:}
  \begin{tabular}{l c c c c c} \toprule
   &GPR &2-stage GP &SLFM &SAGP \\ \midrule
   Poverty rate & 0.599 $\pm$ 0.099 (--) & 0.380 $\pm$ 0.060 (--) & 0.335 $\pm$ 0.052 (2) & {\bf 0.278 $\pm$ 0.032}$^{\star\star}$ (2) \\[1pt]
   Unemployment rate & 0.478 $\pm$ 0.047 (--) & 0.318 $\pm$ 0.032 (--) & 0.278 $\pm$ 0.025 (2 ) & {\bf 0.231 $\pm$ 0.021}$^{\star}$ (2) \\[1pt]
   \bottomrule
  \end{tabular}}
 \end{center}
\end{table}

\end{document}